\documentclass[twoside]{article}

\usepackage[nointegrals]{wasysym}

\usepackage{dcolumn}
\usepackage{amssymb}
\usepackage{caption,tabularx}

\usepackage{array}
\usepackage[table]{xcolor}

\usepackage{algorithm,algpseudocode}
\usepackage{mathtools}
\usepackage{lscape}
\usepackage{graphicx}
\usepackage{url}
\usepackage{multirow}
\usepackage{color}
\usepackage{xspace}
\usepackage{hyperref}
\usepackage{enumerate}
\usepackage{subcaption}
\usepackage{lineno}
\graphicspath{ {images/} }

\makeatletter

\usepackage{ifxetex}
\ifxetex\else
  \usepackage{dblfloatfix}
\fi

\@ifundefined{subparagraph}{
\def\subparagraph{\@startsection{paragraph}{5}{2\parindent}{0ex plus 0.1ex minus 0.1ex}%
{0ex}{\normalfont\small\itshape}}%
}{}

\def\URL#1#2{\@ifundefined{href}{#2}{\href{#1}{#2}}}

\def\UrlOrds{\do\*\do\-\do\~\do\'\do\"\do\-}%
\g@addto@macro{\UrlBreaks}{\UrlOrds}

\makeatother

\usepackage[paperheight=11in,paperwidth=8.3in,margin=2.5cm,headsep=.7cm,top=2.5cm]{geometry}
\usepackage[T1]{fontenc}

\renewenvironment{abstract}
	{\trivlist\item[]\leftskip0pt\par\vskip4pt\noindent
  	\textbf{\abstractname}\mbox{\null}\\}
	{\par\noindent\endtrivlist}

\def\keywords#1{\par\medskip\par\noindent\textbf{Keywords}: #1\par}

 \date{} \emergencystretch 8pt

\makeatletter
\def\author#1{\gdef\@author{\hskip-\tabcolsep%
	\parbox{\textwidth}{\raggedright\bfseries#1\\[1pc]}}}
\def\address[#1]#2{\g@addto@macro\@author{\\\hskip-\tabcolsep\parbox{\textwidth}{\raggedright%
	\normalsize\normalfont\textsuperscript{#1}#2}}}
\let\addresslink\textsuperscript
\def\correspondence#1{\g@addto@macro\@author{\\\hskip-\tabcolsep\parbox{\textwidth}{\raggedright%
	\vspace*{10pt}\normalsize\normalfont~\\#1~\\[12pt]}}}
\def\email#1{\g@addto@macro\@author{\\\hskip-\tabcolsep\parbox{\textwidth}{\raggedright%
	\normalsize\normalfont Emails: #1}}}

\def\title#1{\gdef\@title{\vspace*{-30pt}%
	\raggedright\textbf{\@journaltitle}~\\%
  \raggedright\bfseries\ifx\@articleType\@empty\vspace*{20pt}\else%
  \vspace*{20pt}\@articleType\vspace*{20pt}\\\fi#1}}
\let\@journaltitle\@empty  \def\journaltitle#1{\gdef\@journaltitle{{\normalfont\itshape#1}}}
\let\@articleType\@empty \def\articletype#1{\gdef\@articleType{{\normalfont\itshape#1}}}

\let\@runningHead\@empty \def\RunningHead#1{\gdef\@runningHead{{\normalfont #1}}}

\usepackage{fancyhdr}
\fancypagestyle{headings}{\fancyhf{}
  \fancyhead[R]{\itshape\@runningHead}
  \fancyfoot[C]{\thepage}}
\fancypagestyle{plain}{%
	\fancyhf{}\fancyhead[R]{\LaTeX\ template for Hindawi journal articles}
  \fancyfoot[C]{\thepage}}
\makeatother

\usepackage[%
	numbers,sort&compress%
  ]{natbib}

\usepackage{float,xcolor}

\journaltitle{The Scientific World Journal}
\articletype{Research Article} % Research Article/Review Article/Clinical Study

\begin{document}

% Title of the document
\title{Multi-objective Binary Differential Approach with Parameter Tuning for Discovering Business Process Models: MoD-ProM}

% Author names
\author{%
		A. Sonia Deshmukh\addresslink{1},
  	B. Shikha Gupta\addresslink{2} and
  	C. Naveen Kumar\addresslink{3}
    }
		
% Affiliation
\address[1]{Assistant Professor, Department of Computer Science and Information Technology, KIET Group of Institutions, Delhi-NCR, India}
\address[2]{Associate Professor, Department of Computer Science, S.S. College of Business Studies, University of Delhi, Delhi, India}
\address[3]{Professor, Department of Computer Science, University of Delhi, Delhi, India}

% Corresponding author details
\correspondence{Correspondence should be addressed to 
    	Shikha Gupta: shikhagupta@sscbsdu.ac.in}

% Emails of authors
\email{sonia.cs.du@gmail.com (A. Sonia Deshmukh), shikhagupta@sscbsdu.ac.in (B. Shikha Gupta), nk.cs.du@gmail.com (C. Naveen Kumar)}%

% Running Head
\RunningHead{MoD-ProM}

\maketitle 

% Abstract
\begin{abstract} \label{sec:abstract}
%\textbf{Purpose}-
Process discovery approaches analyze the business data to automatically uncover structured information, known as a process model. The quality of a process model is measured using quality dimensions- completeness (replay fitness), preciseness, simplicity, and generalization. Traditional process discovery algorithms usually output a single process model. A single model may not accurately capture the observed behavior and overfit the training data. We have formed the process discovery problem in a multi-objective framework that yields several candidate solutions for the end user who can pick a suitable model based on the local environmental constraints (possibly varying). We consider the Binary Differential Evolution approach in a multi-objective framework for the task of process discovery. The proposed method employs dichotomous crossover/mutation operators. The parameters are tuned using Grey relational analysis combined with the Taguchi approach. {We have compared the proposed approach with the well-known single-objective algorithms and state-of-the-art multi-objective evolutionary algorithm--- Non-dominated Sorting Genetic Algorithm (NSGA-II).} Additional comparison via computing a weighted average of the quality dimensions is also undertaken. Results show that the proposed algorithm is computationally efficient and produces diversified candidate solutions that score high on the fitness functions. It is shown that the process models generated by the proposed approach are superior to or at least as good as those generated by the state-of-the-art algorithms.\newline

%\textbf{Originality/value}- To the best of our knowledge, the proposed approach is the first attempt to apply the Differential Evolution approach to the task of process discovery as a multi-criterion problem.
%Completeness; Generalization; Simplicity; Preciseness;
\end{abstract}
    
\section{Introduction} \label{sec:Intro}
Processes are ubiquitous in any organization. An efficient organization is built on processes that run in a symphony to achieve growth and customer/employee satisfaction. In the present digital era, organizations maintain the process execution information in the form of transaction logs that are amenable to analyses. However, amidst routine activities, an organization may not analyze the effectiveness of the processes being followed. Process mining aims to extract non-trivial knowledge and exciting insights from data recorded by the information systems, stored in the form of an event log. {In the past decade, process mining adoption has expanded considerably, evidenced by numerous industry and academic use cases, especially in auditing and healthcare, with the field maturing through enhanced tools and techniques \cite{mamudu2024process,HEJAZI2023100188}.} The prominent process mining challenges include process discovery, conformance checking, and enhancement. Process discovery algorithms build a process model from the given event log \citep{van2016process,mining2011discovery,deshmukh2020moea}.  Conformance checking verifies the goodness of the discovered process models. Enhancement techniques extend or improve existing processes by identifying and removing bottlenecks, finding deviations, recommending adjustments, and repairing processes using the information in an event log \citep{van2016process,mining2011discovery}. The present work is focused on the challenge of the process discovery. \par

Process discovery concerns itself with extracting information on existing processes to recognize the bottlenecks, deviations, and inefficiencies in the day-to-day process workflows, providing concrete steps toward business process improvement. The last decade has seen several process discovery techniques that optimize one or more quality metrics, namely, completeness (also known as replay fitness \citep{van2016process}), preciseness, simplicity, and generalization or their weighted function. Typically, process discovery algorithms output a single model. However, a single process model may not always describe the recorded behavior of the log effectively and may be a consequence of over-fitting the training data. \par

In this paper, we present \textbf{M}ulti-\textbf{o}bjective \textbf{D}ifferential approach in \textbf{Pro}cess \textbf{M}ining (MoD-ProM), a process discovery algorithm that generates several competing process models, representing different trade-offs in the quality dimensions. The present work formulates process discovery as a multi-criterion problem. The proposed approach applies the Differential Evolution algorithm and optimizes Completeness and Generalization quality metrics to output several candidate process models. Subsequently, the solutions may either be evaluated by a domain expert to best suit the situation at hand or be chosen by the user based on his/her preference.  \par

The contributions of this proposal are: 
\begin{itemize} \label{contri}
\item A novel application of differential evolution approach for discovering a Pareto-front of the process models.
\item We adapted a binary version of the multi-objective differential evolution algorithm and used dichotomous operators \citep{peng2016dichotomous}.
\item The proposed algorithm (MoD-ProM) is evaluated on ten synthetic and four real-life event logs, and results are compared with the state-of-the-art algorithms.
\item The parameters are tuned using grey relational analysis combined with the Taguchi approach \citep{panda2016multi,lin2004use}.
\item The computation of fitness functions (completeness and generalization) has been reformulated in terms of the causality relation matrix.
\end{itemize}

The results reveal that the proposed approach (MoD-ProM) outperforms the compared algorithms regarding the quality of the process model. Compared to Non-dominated Sorting Genetic Algorithm II (NSGA-II) \citep{deb2002fast}, the proposed algorithm exhibits a lower computational cost. The competing solutions (Pareto set) generated by the proposed approach are better than the non-dominated solutions generated by NSGA-II.\par

%Section 3 gives a brief discussion of the multi-objective differential evolution approach.
The remainder of this paper is organized as follows: section 2 outlines the basic concepts related to process discovery and the related work. Section 3 describes the solution strategy, and section 4 presents the results of the experiments. Finally, section 5 gives the conclusion of the paper. \par

\begin{table}[t]
\caption{An event log with three process instances}
\begin{center}
\scalebox{0.7}{
\begin{tabular}{| p{1.4cm}p{3.0cm}|}
\hline
%\rowcolor{LightCyan} 
Case ID & Process instance\\
\hline
101 & $T_1$ \ $T_2$ \ $T_3$ \\

102 & $T_1$\ $T_2$\ $T_4$\ $T_6$\ $T_5$\ $T_7$\\

103 & $T_1$\ $T_2$\ $T_4$\ $T_5$\ $T_6$\ $T_7$\\
\hline
\end{tabular}}
\end{center}
\label{table:log1}
\end{table}

\section{Background and Related Work} \label{sec_rw}
\subsection{Process Discovery}
Process discovery is an evolving domain that leverages event logs to analyze business processes and present factual insights. An event log is the starting point for process discovery algorithms and represents a business process using the case notation to correlate events. A ‘‘case'' in this notation refers to an instance of a process and is also known as a trace. Each case is assigned a unique ID, called the Case ID. An instance of a process may involve multiple activities or tasks over many days. An occurrence of a task in the context of a particular process instance (case), along with its timestamp, is called an event. Table~\ref{table:log1} gives an example of an event log. In this example, 101, 102, and 103 represent the Case ID of three process instances, and $T_1$, $T_2$, $\ldots$, and $T_7$ represent the various tasks carried out in the system. \par
     
\subsubsection{Visualisation of a Process Model} \label{sec_pm}
A process model can be discovered from the given event log and may be visualized in various forms such as Business Process Modelling Notation (\textit{BPMN}  models), Petri nets, and Data Flow Graphs (\textit{DFGs}), etc. In this paper, the discovered process model is graphically represented as a \textit{Petri net}, a popular method for representation. A Petri net is a bipartite graph, composed of nodes, tokens, and directed arcs. A node could be a place (denoted by a circle) or a transition (denoted by a square). The places and the transitions are joined by directed arcs. For example, in the following figure, $p_1$ and $p_2$ are places and $t_1$ is a transition. \par 

\begin{figure}
%\caption{Nodes and Arcs in a Petri net}
\centering
\includegraphics[width=0.3\textwidth]{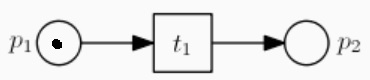}
\label{fig:place1}
\end{figure}

A transition is also called a task. The token is the information that needs to be processed. Each place can hold zero or more tokens.  In the above figure, the place $p_1$ holds a single token. The directed arcs can transfer one token. Transitions cannot store tokens. Arcs connect (input) places to transitions and transitions to (output) places. The state of a Petri net is given by its assignment of tokens to places.\par

A transition is said to be enabled if each input place holds at least one token. In the following figure, $t_1$ transition is enabled. \par

\begin{figure}[H]
%\caption{Enabled transitions} 
\centering
\includegraphics[width=0.3\textwidth]{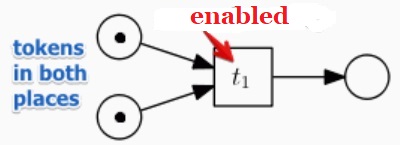}
\label{fig:enable1}
\end{figure}

An enabled transition may fire at any time. When fired, the tokens in the input places are moved to the output places of the transition. Firing of a transition results in a new state of the Petri net. The following figure shows the change in the above Petri net after transition $t_1$ fires. \par

\begin{figure}[H]
%\caption{Firing a transitions}
\centering
\includegraphics[width=0.3\textwidth]{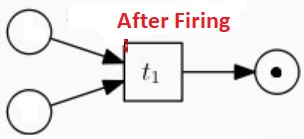}
\label{fig:fire1}
\end{figure}

A transition cannot be enabled if a token is absent (missing token) at any input place. For example, in the following figure, transition $t_1$ cannot be enabled. \par 

\begin{figure}[H]
\centering
\includegraphics[width=0.3\textwidth]{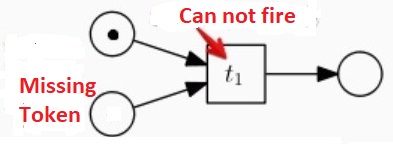}
\label{fig:missing}
\end{figure}

\begin{figure*}[!ht]
\centering
\includegraphics[width=9.5cm,height=3.9cm]{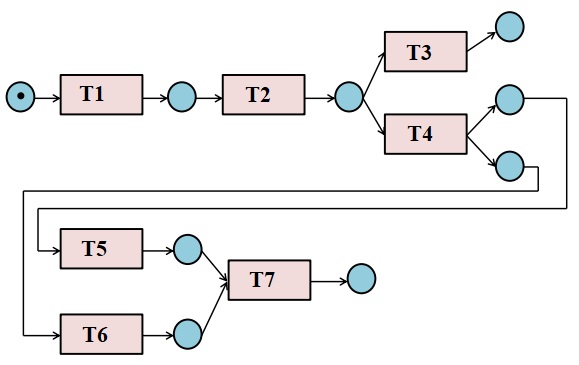}
\caption{Petri net for the Example Event Log of Table~\ref{table:log1} }
\label{fig:SmallDataset}
\end{figure*}

Figure~\ref{fig:SmallDataset} depicts a Petri net that conforms to the example event log in Table ~\ref{table:log1}. \par 

\subsubsection{Well-known Algorithms for Process Model Discovery} \label{sec-algo}
State-of-the-art process discovery techniques include \(\alpha\) \citep{van2004workflow}, \(\alpha\)$^{+}$ \citep{alves2004process}, Multi-phase miner \citep{van2004multi,van2005multi},  Heuristics miner \citep{weijters2006process}, Genetic process mining (GPM) \citep{alves2006genetic}, \(\alpha\)$^{++}$ \citep{wen2007mininga}, \(\alpha\)$^{\#}$ \citep{wen2007miningb}, \(\alpha\)$^{*}$ \citep{li2007process}, Fuzzy miner \citep{van2007finding}, Inductive Logic Programming (ILP) \citep{goedertier2009robust} algorithms \citep{deshmukh2021ga}. Other algorithms in the domain of process discovery include Evolutionary tree miner (ETM) \citep{buijs2012role}, Inductive miner \citep{leemans2013discovering}, Multi-paradigm miner \citep{de2014multi}. \cite{cheng2015hybrid} proposed a hybrid process mining approach that integrates the GPM, particle swarm optimization (PSO), and discrete differential evolution (DE) techniques to extract process models from event logs. \cite{vanden2017fodina} proposed the Fodina algorithm, an extension of the Heuristic miner algorithm \citep{vanden2017fodina}.\par

\cite{buijs2013discovering} proposed an extension of the ETM algorithm \citep{buijs2012role} that discovers a collection of mutually non-dominating process trees using NSGA-II \citep{srinivas1994muiltiobjective}. This algorithm optimizes replay fitness, precision, generalization, simplicity, and the number of low-level edits.\par

\subsubsection{{Motivation for the Proposed Algorithm}} \label{sec:motivation}

{Usually, state-of-the-art process discovery algorithms output a single process model that may overfit the training data. To capture the observed behavior more accurately, we propose a multi-objective algorithm for process discovery. The proposed approach yields several candidate solutions. Subsequently, the solutions may either be evaluated by a domain expert to best suit the situation at hand or be chosen by the user based on the local environmental constraints (possibly varying).} \par 
%have and overcome this challenge the concept of multi-objective algorithms came. Due to the higher computational time of NSGA-II, we have formed the process discovery problem in a multi-objective framework that yields several candidate solutions for the end user who can pick a suitable model based on the local environmental constraints (possibly varying) with lower computational time \cite{HEJAZI2023100188}.}
{The proposed algorithm formulates the problem of process model discovery in a multi-objective framework using the Differential evolution approach \citep{storn1997differential}. Differential evolution (DE) is a versatile and stable evolutionary algorithm. It evolves individuals through the perturbation of population members with scaled differences of distinct population members. DE algorithm has consistent robust performance and is suitable for solving various numerical optimization problems \citep{huang2013improved}.} \par

\subsection{Multi-objective Binary Differential Evolution}
The proposed algorithm employs a binary version of the differential evolution approach to suit the process mining domain. While DE was initially designed to operate in continuous space, \cite{peng2016dichotomous} proposed a Binary DE (BDE) algorithm based on dichotomous mutation and crossover operators. The authors \citep{peng2016dichotomous} verified that compared to other notable BDE variants, the dichotomous BDE improves the diversity of the population and enhances the exploration ability in the binary search space. Also, it has been shown that as compared to other BDE variants, the dichotomous algorithm does not involve any additional computation cost and is faster than other variants of BDE \citep{peng2016dichotomous}.\par

The past decade has seen the application of the DE approach to problems where the optimization of multiple objectives is required. \cite{robict2005demo} first proposed a DE-based approach for multi-objective real-coded optimization problems. According to \cite{tuvsar2007differential}, in the case of binary-coded optimization problems, multi-objective BDE algorithms explore the decision space more efficiently than other multi-objective evolutionary algorithms. Subsequently, multi-objective BDE algorithms were also proposed \cite{li2013non}, \cite{xue2014differential}, \cite{sikdar2015mode}, \cite{bidgoli2019novel}, \cite{zhang2020binary}. \par

\section{Materials and Methods} \label{sec:proposedApproach}
%Proposed Multi-objective Differential Approach for Process Mining (MoD-ProM)
A process discovery algorithm is a function that maps an event log onto a process model that best represents the behavior seen in the event log. In the present work, a process model is represented by a causality relation matrix $C$= ($c_{t_1,t_2}$), where $t_1,\ t_2\ \in [1, n]$ represent the tasks, $c_{t_1,t_2} \in \{0,1\}$, and n is the number of tasks in the given event log. That is, an individual in the population is binary-coded. We, therefore, adapted a binary version of the multi-objective differential evolution algorithm using dichotomous operators \citep{peng2016dichotomous}. The steps for the proposed multi-objective differential approach for process mining (MoD-ProM) are outlined in Algorithm \ref{alg:DEMO}. These steps are explained in the following subsections. \par
 
\subsection{Initialization} \label{sec:initialization}
The given event log E with n tasks is first consolidated into a dependency measure matrix $D$ indicating the degree of dependencies between tasks \citep{alves2006genetic}. Considering the example in Figure~\ref{fig:SmallDataset}, where $T_1$, $T_2$, $\ldots$, $T_7$ represent the tasks. A dependency exists between activities $T_1$ and $T_2$ if, in a trace,  either $T_1$ directly precedes $T_2$ or vice versa. This is indicated by the presence of either the strings $T_1\ T_2$ or $T_2\ T_1$ in a process instance (trace) of the event log. The strength of dependency is proportional to the frequency of occurrence of these strings. In the example log (Figure ~\ref{fig:SmallDataset}), $T_1$ directly precedes $T_2$ whereas the string $T_2\ T_1$ does not occur at all. That is, in the given system, task $T_1$ is more likely to be the cause of task $T_2$ than vice versa. Dependency measure is computed by counting the length-one-loops (for example, $T_1\ T_2$), self-loops (for example, $T_1\ T_1$), length-two-loops (for example, $T_1\ T_2\ T_1$), and parallel tasks (for example, $T_1\ T_2$ and $T_2\ T_1$ occur an equal number of times). In the example log (Figure ~\ref{fig:SmallDataset}), $T_5$ and $T_6$ are parallel tasks and $T_1\ T_2$ is a length-one-loop \citep{gupta2024mantaray}. \par

As proposed by \citep{de2007genetic}, the present work represents a process model as a causality relation matrix. To represent a process model, \cite{de2007genetic} have favored a causality relation matrix over the more popular Petri net since representing the population individual as a causality relation matrix makes it easier to initialize the population and define the genetic operators. While a causality relation matrix can be directly derived from the information in the event log, in Petri nets, there are places whose existence cannot be derived directly from the event log \citep{de2007genetic}. The mapping between a Petri net and a causality relation matrix is detailed in \citep{de2007genetic}. We have graphically depicted both representations for an example event log in Section \ref{sec:processModelRep} titled "Process Model Representation". \par

\subsection{{Objective Functions and Fitness Evaluation}} \label{FE} 
{In the proposed algorithm, we use a novel combination of completeness and generalization as objective functions. Completeness is an important quality dimension because a discovered process model is expected to describe the behavior stored in the log. Completeness is the process of computation of all the parsed tasks while replaying the traces of the log in the model. The missing tokens in a trace and the extra ones left behind during parsing (unconsumed tokens) contribute to the penalty value. Generalization shows whether the process model accurately represents the system as it is and is not "overfitting" to the behavior observed in the event log \citep{buijs2014quality}. Completeness \citep{alves2006genetic} and generalization \citep{van2013alignment} are computed as in Algorithms \ref{alg:comp} and \ref{alg:gen} respectively. The algorithms make use of the following function for a given event log E:} \par

\begin{equation}
\raggedleft
follows(t_1, t_2, E) = 
 \begin{cases}
 ${ 1  if $t_1$ $t_2$ is length-one-loop in E}$\\
 ${ 0,   otherwise}$\\
 \end{cases}
 \label{equ:causality2}
 \end{equation}
 
 \begin{equation}
 \raggedleft
follows_k(t_1, t_2, E) = 
 \begin{cases}
 ${ 1  if $t_2$ is the $k^{th}$ task after $t_1$ in E}$\\
 ${ 0,   otherwise}$\\
 \end{cases}
 \label{equ:causality1}
 \end{equation}

\begin{algorithm}
\caption{Completeness (C, E)}
\begin{algorithmic}[1]
\State $\texttt{n= Total number of tasks/activities in the Event Log E}$
\State $\texttt{\#Traces= Total number of traces in E}$		
\State $\texttt{ParsedTasksRatio =$\frac{1}{n}$ $\sum\limits_{t_1=1}^{n} \sum\limits_{t_2=1}^{n}c_{t_1,t_2}$ $* follows(t_1, t_2, E)$}$
\State $\texttt{\#MissingTokens = $\sum \sum(1 - c_{t_1,t_2})$* $follows(t_1, t_2, E)$}$
\State $\texttt{\#$Traces_{MissingTokens}$ = $|$traces of MissingTokens$|$}$
\State $\texttt{\#TracesWithoutMissingTokens = \#Traces - \#$Traces_{MissingTokens}$  + 1}$
\State $\texttt{\#ExtraTokens1 = {$\sum\limits_{t_2=1}^{n} ( \prod\limits_{t_1} c_{t_1,t_2})$ where $t_1 \in [1, n]$ such that $\sum\limits_{t_2=1}^{n} c_{t_1,t_2} \ge 1$ }}$
\State $\texttt{\#ExtraTokens2 =$\sum\limits_{\forall(t_1, t_2, t_3)} [ c_{t_2,t_3} * c_{t_1,t_3}* (1- c_{t_1,t_2})* follows(t_1, t_2, E)$ }$
${ \ \ \ \ \ \ \   \ \ \ \ \ \ \ \ \ \ \ \ \ \ \ \ \ \ \ \ \ \ \ \ \ \ \ \ \ \   * follows_k(t_1, t_3, E) *follows_{k-1}(t_2, t_3, E) ] }$
\State $\texttt{\#ExtraTokens= \#ExtraTokens1 + \#ExtraTokens2}$
\State $\texttt{\#$Traces_{ExtraTokens}$ = $|$traces of ExtraTokens$|$}$
\State $\texttt{\#TracesWithoutExtraTokens = \#Traces - $Traces_{ExtraTokens}$ + 1}$  
\State $\texttt{Penalty = $\frac{ 1}{n}\left[ {\dfrac{\#MissingTokens}{\#TracesWithoutMissingTokens} + \dfrac{\#ExtraTokens}{\#TracesWithoutExtraTokens}} \right]$}$  
		\State $\texttt{Completeness = ParsedTasksRatio - Penalty}$    		    
		\State $\textit{Return Completeness}$	
	\end{algorithmic}
		\label{alg:comp}
		\end{algorithm}

 \begin{algorithm}
		\caption{Generalization Computation: Generalization (C, E)}
			\begin{algorithmic}[1]
			%\State $\textit{$follows(t_1, t_2, E)$ = 1 if in a trace, $t_2$ follows $t_1$}$
			\State $\texttt{n= number of tasks/activities in E}$		
		 \State $\texttt{$executions(t_1, t_2)$ = $ |(follows(t_1, t_2))|$ * $c_{t_1, t_2}$, $t_1,\ t_2\ \in [1, n]$ }$               
	\State $\texttt{$executionsFrequency(t_1)$ = $\sum\limits_{t_2=1}^{n} {executions(t_1, t_2)}$, $t_1\ \in [1, n]$ }$		    
\State $\texttt{$executionsInvSqrt(t_1)=\frac{1}{ \sqrt{executionsFrequency(t_1)}}$, $t_1\ \in [1, n]$}$
 \State $\texttt{$Generalization = 1- \frac{1}{n} {\sum\limits_{t_1=1}^{n}{executionsInvSqrt(t_1)}}$}$ 
	\State $\texttt{Return Generalization}$
		\end{algorithmic}
		\label{alg:gen}
		\end{algorithm}

{The present proposal performs an additional analysis of the discovered process models by evaluating their preciseness and simplicity values. The preciseness value of a model is relative to an event log and quantifies the behavior existing in the model but not observed in the event log\citep{van2016unified}. A process model with a high precision value is not expected to show behavior not observed in the event log \citep{mannhardt2016measuring}. Completeness and preciseness only consider the relationship between the event log and the process model. However, just a portion of all potential behavior that the system permits is recorded in the event log. Simplicity, instead of telling about the behavior observed in the event log, shows the internal structure of the discovered model. Preciseness \citep{buijs2012role} and simplicity \citep{vazquez2015prodigen} values are computed as in Algorithms \ref{alg:pre} and \ref{alg:sim} respectively.} \par

 \begin{algorithm}
\caption{Preciseness (C, E)}
\begin{algorithmic}[1]
				%\State $\textit{$follows(t_1, t_2, E)$ = 1 if in a trace, $t_2$ follows $t_1$}$
\State $\texttt{n= number of tasks/activities in E}$		
 \State $\texttt{$\#visits(t_1)$ = $|a_{t_1}| $, $a_{t_1}$ represent an activity in E, $t_1\in [1, n]$ }$               
\State $\texttt{$\#outgoingEdges(t_1)$ = $\sum\limits_{t_2=1}^{n} (follows(t_1, t_2) > 0)$, $t_1\in [1, n]$ }$		    
\State $\texttt{$\#usedEdges(t_1)$=$\sum\limits_{t_2=1}^{n} follows(t_1, t_2) * c_{t_1, t_2}$, $t_1\in [1, n]$}$
\State $\texttt{$\#visitedMarkings =\sum  \#visits(t_1) * \frac{\#outgoingEdges(t_1)- \#usedEdges(t_1)}{\#outgoingEdges(t_1)}$}$
	\State $\texttt{$ \#totalMarking =\sum \#visits(t_1) * \#outgoingEdges(t_1)$}$
	 \State $\texttt{$Preciseness = 1- \frac{\#visitedMarkings}{\#totalMarking}$}$ 
		\State $\texttt{Return Preciseness}$
			\end{algorithmic}
			\label{alg:pre}
			\end{algorithm}

 \begin{algorithm}
\caption{Simplicity (C, E)}
\begin{algorithmic}[1]
\State $\texttt{n= number of tasks/activities in E}$			               
\State $\texttt{$sim(t_1)$ = $\sum\limits_{t_2=1}^{n} c_{t_1, t_2}$ + $\sum\limits_{t_1=1}^{n} c_{t_1, t_2}$, $t_1 \in [1, n]$ }$		    
\State $\texttt{$cardinality=\sum sim(t_1)$}$
 \State $\texttt{$Simplicity =  \frac{1}{cardinality}$}$
\State $\texttt{Return Simplicity}$
\end{algorithmic}
\label{alg:sim}
\end{algorithm}

\subsection{{Constraints and Decision Variables}} \label{sec:constraints}
For a given event log E, Dependency measure matrix $D= (D(t_1, t_2))$ is used to generate causality relation matrices $C^i$= ($c_{t_1,t_2}^{i}$) where $i \in [1, N]$ , $t_1,\ t_2\ \in [1, n]$, N is the population size (Algorithm \ref{alg:algorithm2}). {The dependency measure matrix and the causality matrix correspondingly represent the constraints and the decision variables for the problem.} Each causality relation matrix represents an individual of the initial population and is computed as \citep{alves2006genetic}: \par

  \begin{equation}
 c_{t_1,t_2}^i = 
 \begin{cases}
 ${ 1 if r $<$ $D(t_1,t_2)$}$\\
 ${ 0,   otherwise}$\\
 \end{cases}
 \label{equ:causality}
 \end{equation}

$r \in [0,1)$ is a random number.  \par

\begin{algorithm}
\caption{Computation of initial population of N individuals, each of size n $\times$ n: Initialize Population (n, N, D)}
\begin{algorithmic}[1]
\For{\texttt{i=1:N}}
   \State $\texttt{$C^i$=zeros(n,n)}$   
 	\For{every tuple $(t_1, t_2)$ $\in$ [1, n] $\times$ [1, n]}
 	        \State $\texttt{ Generate a random number r}$
 	        \State $\texttt{$c_{t_1,t_2}^i \gets$ r $<$ $D(t_1,t_2)$}$	
 	 \EndFor
\EndFor
\State $\texttt{ Return Pop = \{$C^1$, $C^2$, $\ldots$, $C^N$\}}$
\end{algorithmic}
\label{alg:algorithm2}
\end{algorithm}

\subsection{Mutation}
For a population member $C^{i}$= ($c_{t_1,t_2}^i$), $i \in [1, N]$ , $t_1,\ t_2\ \in [1, n]$, two other causal matrices $C^{r_1}$, $C^{r_2}$, $r_1$ $\neq$ $r_2$ $\neq$ i, $r_1, r_2 \ \in [1, N]$ are chosen randomly from the current population. A mutant individual $V^{i}$= ($v_{t_1,t_2}^i$) is then created using the following dichotomous mutation scheme \citep{peng2016dichotomous}. \par

\begin{equation}
\begin{multlined}
v_{t_1,t_2}^i = ((c_{t_1,t_2}^{r_1} \oplus c_{t_1,t_2}^{r_2})\land rand)  
 \lor (\neg (c_{t_1,t_2}^{r_1}\oplus c_{t_1,t_2}^{r_2})\land c_{t_1,t_2}^{r_1})
\end{multlined}
\label{equ:mutation1}
\end{equation}

where rand $\in$ $\{0,1\}$, $\land$ denotes the AND operator,  $\lor$ denotes the OR operator, $\neg$ denotes the NOT operator, and $\oplus$ denotes the XOR operator. Equation~\ref{equ:mutation1} can also be expressed as: \par

\begin{equation}
v_{t1,t2}^i=
 \begin{cases}
 ${ rand  \  if ${c_{t_1,t_2}^{r_1} \oplus c_{t_1,_t2}^{r_2}}$ $=$ 1}$\\
 ${ $c_{t_1,t_2}^{r_1}$  \   if ${c_{t_1,t_2}^{r_1} \oplus c_{t_1,t_2}^{r_2}}$ $=$ 0}$\\
 \end{cases}
 \label{equ:mutation2}
\end{equation}

That is, if $c_{t_1,t_2}^{r_1}$ and $c_{t_1,t_2}^{r_2}$ are distinct, then the corresponding bit of the mutant individual $v_{t_1,t_2}^{i}$  is randomly chosen as “0” or “1”; otherwise, $v_{t_1,t_2}^{i}$ is set as $c_{t_1,t_2}^{r_1}$. \par

\subsection{Crossover}
The Dichotomous crossover operator \citep{peng2016dichotomous} starts from the mutant individual $V^{i}$= $v_{t_1,t_2}^i$, obtained after application of the dichotomous mutation operator. In this step, the original individual $C^{i}$= ($c_{t_1,t_2}^i$) and the mutated individual $V^{i}$ are used to generate a candidate individual $U^{i}$= $u_{t_1,t_2}^i$ using the following equation: \par

 \begin{equation}
 u_{t_1,t_2}^{i} = 
 \begin{cases}
 ${ $v_{t_1,t_2}^i $        if $rand_{t_1,t_2}$ $<$ $CR_{t_1,t_2}$}$\\
 ${$c_{t_1,t_2}^{i}$,   otherwise}$\\
 \end{cases}
 \label{equ:CR1}
 \end{equation} 
   where $rand_{t_1,t_2}$ $\in$ [0, 1], 
     \begin{equation}
CR_{t_1,t_2} = 
 \begin{cases}
 ${ $CR_1 $ if ($c_{t_1,t_2}^{r_1}$ $\oplus$ $c_{t_1,t_2}^{r_2}$)  $=$ 0    }$\\
 ${ $CR_2$    if ($c_{t_1,t_2}^{r_1}$ $\oplus$ $c_{t_1,t_2}^{r_2}$) $=$ 1}$\\
 \end{cases}
 \label{equ:CR2}
 \end{equation} 
 
 This operation uses two crossover probabilities $CR_1$ and $CR_2$ based on dichotomous psychological thinking or "black and white'' thinking, with a proclivity for only seeing extremes. After mutation, to generate a candidate individual, if the bits in the randomly chosen individuals from the original population are the same (distinct), then crossover probability $CR_1$ ($CR_2$) is used. This approach induces diversity in the population and enhances the exploration ability of the proposed approach \citep{peng2016dichotomous}. \par
  		
 \subsection{Selection}
 In this section, we outline the selection procedure (Algorithm \ref{alg:css}) used to determine the individuals to be preserved from the current population Pop= \{$C^{1}$, $C^{2}$,\ldots, $C^{N}$\}, and the candidate population $Pop_U$= \{$U^{1}$, $U^{2}$,\ldots, $U^{N}$\} generated after the crossover operation. The process involves identifying the non-dominated individuals.
 \begin{algorithm}
\caption{Selection of the fittest individuals from current (Pop) and candidate ($Pop_U$) populations, each of size N: Selection ($Pop_U$, Pop)}
\begin{algorithmic}[1]
%\State $\texttt {Selection($U$, $C$)}$
\For{\texttt{i = 1 to N}}
\If{ $U^{i}$ $\prec$ $C^{i}$ \  (Equation \ref{equ:se})}
     \State $\texttt {Pop}\gets\textit {($Pop - C^{i}$) $\cup$ $U^{i}$ }$
  \ElsIf{$C^{i}$ $\prec$ $U^{i}$}
  \State $\texttt {Pop}\gets\textit {Pop}$
  \Else
   \State $\texttt {Pop}\gets\textit {$Pop \cup U^{i}$}$
    \EndIf
   \State$\texttt{Return Pop}$ 
\EndFor
\end{algorithmic}
\label{alg:css}
\end{algorithm}

The $i^{th}$ individual from the current population (parent) ($C^{i}$) is said to dominate ($\prec$) the corresponding $i^{th}$ individual in the candidate population (child) ($U^{i}$) if the parent is superior for both the objectives of completeness and generalization, that is, \par

\begin{equation} 
 C^{i} \prec U^{i} = 
 \begin{cases}
 ${ 1 \ if $f_c(U^{i}) \geq  f_c(C^{i})$ \ \&\&\  $f_g(U^{i}) \geq  f_g(C^{i})$  }$\\
 ${ 0,   otherwise}$\\
 \end{cases}
\label{equ:se}
 \end{equation}
 
 where $f_c$ and $f_g$ denote the completeness and generalization values respectively. \par 
 
 If the parent (child) dominates the child (parent), then the parent (child) is preserved while the child (parent) is discarded. When neither parent nor child is superior to each other, both the parent and the child are retained.\par 
After eliminating dominated individuals, the number of remaining non-dominated individuals will be between N and 2*N.  Since the population size to be carried for the next generation is N, a truncation procedure based on non-dominated sorting (Algorithm \ref{alg:NDS}) and crowding distance (Algorithm \ref{alg:cdm}) is applied \citep{robict2005demo}. \par 

Non-dominated sorting algorithm (Algorithm \ref{alg:NDS}), involves finding rank $1$ individuals of the population that are not dominated by any other individual. Rank $2$ is assigned to those individuals of the population that are dominated by rank $1$ individuals, and so on. \par 

\begin{algorithm}
\caption{Non-Dominated Sorting Algorithm: NonDominatedSorting (Pop, N)}
\begin{algorithmic}[1]
%\State $\textit{}$
%\State $\texttt{Input: population C$=$ $C^i$, $i \in [1, N]$}$ 
\For{\texttt{r = 1 to N}}
\State $\texttt{G$_{r}$ $=$ $\emptyset$ ( Set of individuals dominated by Pop$^{r}$)}$
\State $\texttt{m$_{r}$ = 0 (number of individuals that dominate Pop$^{r}$)}$ 
\For{\texttt{s = 1 to N}}
	\If {(Pop$^{r}$ $\prec$Pop$^{s}$) (Equation \ref{equ:se}) }
       	\State $\texttt{G$_{r}$ = G$_{r}$ $\cup$ \{Pop$^{s}$\} (Add Pop$^{s}$ into G$_{r}$) }$
       	\ElsIf {(Pop$^{s}$ $\prec$ Pop$^{r}$) }
		\State $\texttt{m$_{r}$ = m$_{r}$+1 \ (increment domination counter m$_{r}$)}$	
	\EndIf
	\EndFor
\If {m$_{r}$= 0 (no solution dominates Pop$^{r}$)  }	
\State $\texttt{Front$_{1}$= Front$_{1}$ $\cup$ \{Pop$^{r}$\} (Pop$^{r}$ is a member of first front)}$
\EndIf
	\EndFor
	
	\State $\texttt{Initialize m$_{z}$ = $|$G$_{r}$$|$ }$ 
\State $\texttt{k = 1 (initialize front counter)}$ 
\While {\texttt{Front$_{k}$ $\neq$ $\emptyset$}}
\State $\texttt{G$_{z} $=$ $ $ \emptyset$ (Store individuals of next front)}$
\For{\texttt{r = 1 to $|$Front$_{k}$$|$}} %(for each member C$^{r}$ in Front$_{k}$)}}
\For{\texttt{z = 1 to  $|$G$_{r}$$|$}}   %(modify each member from the set G$_{r}$)}}
\State $\texttt{m$_{z}$ = m$_{z}$-1 (decrement m$_{z}$ by one)}$
\If {m$_{z}$= 0 }
	\State $\texttt{G$_{z}$= G$_{z}$ $\cup$ \{Pop$^{z}$\}(Pop$^{z}$ is a member of next front G$_{z}$)}$    
	\EndIf\EndFor \EndFor
\State $\texttt{k = k+1}$ 
\State $\texttt{Front$_{k}$ = G$_{z}$ (Next front is formed with all members of G$_{z}$)}$ 
\EndWhile

\State$\texttt{Return Front$_{1}$, Front$_{2}$,\ldots, Front$_{N}$}$
\end{algorithmic}
\label{alg:NDS}
\end{algorithm}

%\subsubsection{Crowding Distance}
If the number of non-dominated solutions is greater than the population size N, Euclidean distance is used to truncate individuals from the most crowded region (Algorithm \ref{alg:cdm}). If the rank 1 individuals are less than N, then rank 2 individuals are added, and so on. 
 
\begin{algorithm}
\caption{ Pseudo-code for Crowding Distance Algorithm: CrowdingDistance $(C_s, E)$, $C_s$ : Non-dominated solutions}
\begin{algorithmic}[1]
%\State $\textit{}$
\State $\texttt{ l = $|C_s|$ // number of non-dominated solutions}$
\State $\texttt{Initialise the distance vector d= (d$_{1}$,d$_{2}$,...,d$_{l}$)=(0,0,...,0)}$
\For{\texttt{j= 1 to l}}
\State $\texttt{Compute $f_{1}^j$ = Completeness ($C_s^j$, E)  (Algorithm \ref{alg:comp})}$
\State $\texttt{Compute $f_{2}^j$ = Generalization ($C_s^j$, E) (Algorithm \ref{alg:gen})}$
\EndFor
\For{\texttt{j= 1 to 2}}
\State $\texttt{Sort $\{f_{j}^1,\ldots, f_{j}^l\} $ in descending order}$
\State $\texttt{d$_{1}$ = d$_{l}$ = $\infty$,  // distance associated with best and worst point}$
\For{\texttt{i= 2 to l-1}}
\begin{equation*}
d_{i}=d_{i}+ \dfrac {|f_{j}^{i+1}-f_{j}^{i-1}|} {|f_{j}^{1}-f_{j}^{l}|}
\end{equation*}
\EndFor
\EndFor
\State $\texttt{Remove individual having smallest d value}$
\State $\texttt{Return Pareto solutions with uniform distribution}$
% and good diversity}$
% $\texttt{by eliminating individuals based on crowding distance d}$
\end{algorithmic}
\label{alg:cdm}
\end{algorithm}

\begin{algorithm*}
\caption{Pseudo code for Multi-objective Differential Approach for Process Mining (MoD-ProM)}
\label{euclid}
\begin{algorithmic}[1]
\State $\texttt {Input: N: the population size}$
\State $\texttt {Read the given event log E, initialize n = \#tasks in E}$
\State $\texttt {Calculate dependency relations (D) between n tasks}$
 $\texttt{\citep{alves2006genetic}}$
\State $\texttt {Pop = Initialize Population (n, N, D) (Algorithm \ref{alg:algorithm2})}$
$\texttt{Pop = \{$C^1$, $C^2$, $\ldots$, $C^N$\}}$
\For{\texttt{i= 1 to N}}
\State $\texttt{Compute $f_{C}^i$ = Completeness ($C^i$, E)  (Algorithm \ref{alg:comp})}$
\State $\texttt{Compute $f_{G}^i$ = Generalization ($C^i$, E) (Algorithm \ref{alg:gen})}$
\EndFor
%\State $\texttt {Let g = 1. //generation number}$
\While {Stopping criteria (maximum 100 iterations or no improvement for 10 iterations) are not met}
 	\For{\texttt{i = 1 to N }}
 	\State $\texttt {Randomly select two other individuals $C^{r_1}$, $C^{r_2}$, where $r_1$ $\neq$ $r_2$ $\neq$ i}$
\State $\texttt {Generate mutant individual $V^{i}$ using dichotomous mutation operator (Equation~\ref{equ:mutation1},~\ref{equ:mutation2})}$
\State $\texttt {Generate candidate individual $U^{i}$ using dichotomous crossover operator (Equation~\ref{equ:CR1},~\ref{equ:CR2})}$
\State $\texttt {Compute completeness \& generalization of $U^{i}$}$

 \EndFor 
\State $\texttt {$Pop_U$ = $\cup_i$   $U^{i}$}$
\State $\texttt {Pop= Selection ($Pop_U$, Pop) (Algorithm \ref{alg:css}) }$
%     \State $\texttt {If $C^{i}(g)$ $\prec$ $U^{i}(g)$, $U^{i}(g)$ is discarded and $C^{i}(g)$ is preserved.}$
%     \State $\texttt {Otherwise, both are retained in the population (Algorithm \ref{alg:css})}$ 
       
\If {Size of (Pop) $>$ N}
	\State $\texttt {$C_s$ = NonDominatedSorting (Pop, N) (Algorithm \ref{alg:NDS})}$
	\State $\texttt{Pop = CrowdingDistance ($C_s$, E) (Algorithm \ref{alg:cdm})}$	  	
	\EndIf  
	  \EndWhile 
\State $\texttt {Return the Pareto-optimal solutions (Pop)}$
\end{algorithmic}
\label{alg:DEMO}
\end{algorithm*} 

\section{Results and Discussion} 
\subsection{Experimentation} \label{sec:experiment}
The proposed algorithm is tested on both synthetic and real-world datasets (Table~\ref{table:ExperimentalDataset1}). Over the last decade, BPI challenge event logs have become important real-world benchmarks in the data-driven research area of process mining. The proposed algorithm is tested for three BPI event logs, namely, BPI 2012 \citep{van2012event}, BPI 2013 \citep{steeman2013bpi} and BPI 2018 \cite{challenge20184tu}, varying in the number of tasks, number of traces, and their domain. BPI 2012 is one of the most studied datasets in process mining. This dataset contains 13,087 traces, and 23 tasks and is derived from a structured real-life loan application procedure released to the community by a Dutch financial institute. The BPI 2013 dataset is from the IT incident management system of Volvo Belgium with 7554 traces and 13 tasks. BPI 2018 covers the handling of applications for EU direct payments for German farmers from the European Agricultural Guarantee Fund. BPI 2018-reference dataset contains 43802 traces and 6 tasks. The proposed algorithm is also tested on a real-life medical event log containing events of sepsis cases from a hospital with 1000 traces and 16 tasks \citep{mannhardt10sepsis}. The proposed algorithm is also run for synthetic logs (ETM, g2-g10 \citep{alves2006genetic,vazquez2015prodigen,gupta2024mantaray}). \par  
 
The proposed approach is compared with state-of-the-art algorithms, \(\alpha\)$^{++}$  \citep{wen2007mininga}, Heuristic Miner \citep{weijters2006process}, Genetic Miner \citep{alves2006genetic}, ILP  \citep{goedertier2009robust} and Inductive Miner  \citep{leemans2013discovering} algorithms. For the compared algorithms, the completeness, preciseness, and simplicity values for the synthetic datasets are taken as reported by \citep{vazquez2015prodigen}.  However, \cite{vazquez2015prodigen} does not report the value of generalization for these datasets. For the models generated using the Prom tool, \(\alpha\)$^{++}$, Heuristic Miner, Genetic Miner, and ILP algorithms, the generalization value is computed using the Cobefra tool \citep{vanden2013comprehensive,vazquez2015prodigen}. We have also compared the proposed strategy with the NSGA-II algorithm for process discovery. \par 

In the proposed multi-objective differential approach for process mining (MoD-ProM), the population size is set to 100, and the value of control parameters $CR_1$ and $CR_2$ is tuned using grey relational analysis combined with the Taguchi approach (Section~\ref{par}). \par 

The algorithm is run for a maximum of 100 iterations as the proposed algorithm converges before 100 iterations for most datasets. The total number of runs is fixed at 30. \par

\begin{table}[t]
\caption{Details of Datasets Used in Experiments}
\begin{center}
\scalebox{0.7}{
\begin{tabular}{| p{1.9cm}| p{3.4cm}|p{1.5cm}| p{1.0cm}|p{1.5cm}|p{5.5cm}| }
\hline 
Type &Event-log  & Tasks & Traces & Events&Source \\ 
\hline
\multirow{10}{*}{\begin{minipage}{0.5in}\textbf{Synthetic Event logs}\end{minipage}}&ETM & 7 & 100 &790 &\multirow{10}{*}{\begin{minipage}{0.5in}\textbf{\citep{alves2006genetic,vazquez2015prodigen}}\end{minipage}}\\

&g2 & 22 & 300 & 4501 & \\

&g3 & 29 & 300 & 14599& \\

&g4 & 29 & 300 & 5975& \\

&g5 & 20 & 300 & 6172& \\

&g6 & 23 & 300 & 5419& \\

&g7 & 29 & 300 & 14451 &\\

&g8 & 30 & 300 & 5133& \\

&g9 & 26 & 300 & 5679& \\

&g10 & 23 & 300 & 4117& \\
\hline
\multirow{4}{*}{\begin{minipage}{0.5in}\textbf{Real-life Event logs}\end{minipage}}&BPI 2012  & 23 & 13087 & 262200&\citep{van2012event} \\
& BPI 2013-incident & 13 & 7554 &  65533 &\citep{steeman2013bpi} \\
&BPI 2018-reference & 6 & 43802 & 128554& \citep{challenge20184tu}\\
&Sepsis&16&1050&150000&\citep{mannhardt10sepsis}\\
\hline
\end{tabular}}
\end{center}
\label{table:ExperimentalDataset1}
\end{table}

\begin{figure*}[ht!]
\captionsetup{font=footnotesize}
           \centering
\subcaptionbox{}{\includegraphics[width=0.48\textwidth,height=6.4cm]{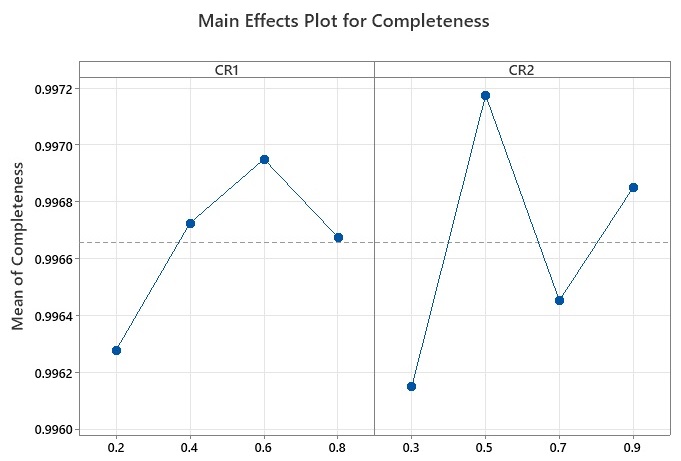}}\label{figure:O:1:ga-1}%
\hfill % <-- Seperation
\subcaptionbox{}{\includegraphics[width=0.49\textwidth,height=6.4cm]{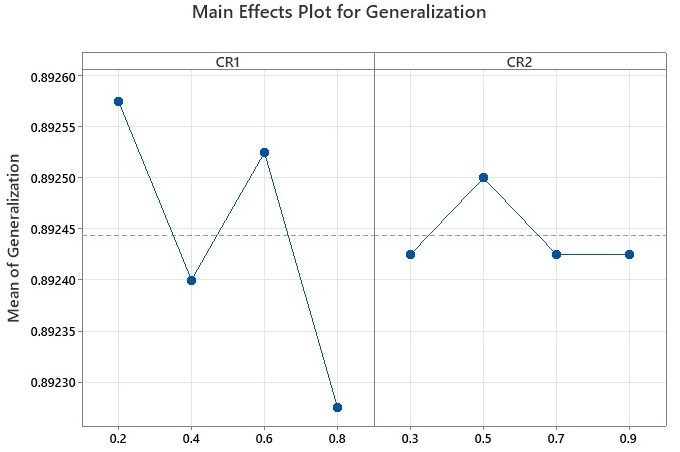}} \label{figure:O:1:gb-2}%
\caption{Main effects plot of Completeness and Generalization}
 \label{figure:O:1:gg1}
        \end{figure*}

%\begin{sidewaystable}
\begin{table}
    \centering
\caption{Experimental results for Completeness and Generalization as per Taguchi L16 Orthogonal Array}
\scalebox{0.7}{
\begin{tabular}{|p{1.6cm}p{1.4cm}p{1.4cm} p{1.9cm} p{2.1cm}|  }
\hline
\multirow{1}{*}{\begin{minipage}{0.5in} RunNo.\end{minipage}}&\multicolumn{2}{c}{\multirow{1}{*}{Process Parameters}}&\multicolumn{2}{c|}{Experimental Results}\\
\cline{2-5}
&	CR1&	CR2&	Completeness&	Generalization\\
\hline
1&	0.2&	0.3&	0.9959&	0.8931\\
2&	0.2&	0.5&	0.9996&	0.8927\\
3&	0.2&	0.7&	0.99531&	0.8926\\
4&	0.2&	0.9&	0.9943&	0.8919\\
5&	0.4&	0.3&	0.9945&	0.8917\\
6&	0.4&	0.5&	0.998&	0.8928\\
7&	0.4&	0.7&	0.9971&	0.8922\\
8&	0.4&	0.9&	0.9973&	0.8929\\
9&	0.6&	0.3&	0.9958&	0.892\\
10&	0.6&	0.5&	0.9961&	0.893\\
11&	0.6&	0.7&	0.9994&	0.8926\\
12&	0.6&	0.9&	0.9965&	0.8925\\
13&	0.8&	0.3&	0.9984&	0.8929\\
14&	0.8&	0.5&	0.995&	0.8915\\
15&	0.8&	0.7&	0.994&	0.8923\\
16&	0.8&	0.9&	0.9993&	0.8924\\
\hline
\end{tabular}}
\label{table:orthogonaltable}
%\end{sidewaystable}
\end{table}

\begin{table}
    \centering
\caption{Grey Relational Generation Values}
\scalebox{0.7}{
\begin{tabular}{|p{1.6cm} p{1.9cm} p{2.1cm}|  }
\hline
Run No.& Completeness&	Generalization\\
\hline
1&	0.339 &	1.000\\
2	&1.000 &	0.750\\
3&	0.234 &	0.687\\
4	&0.054 &	0.250\\
5&0.089 &	0.125\\
6	& 0.714 &	0.813\\
7 &	0.554	&0.438\\
8 &	0.589 &	0.875\\
9 &	0.321 &	0.313\\
10 &	0.375 &	0.938\\
11 &	0.964 &	0.687\\
12 &	0.446 &	0.625\\
13	& 0.786 &	0.875\\
14 &	0.179 &	0.000\\
15 &	0.000 &	0.500\\
16 &	0.946 &	0.562\\
\hline
\end{tabular}}
\label{table:grey1}
%\end{sidewaystable}
\end{table}

\begin{table}
    \centering
\caption{Grey Relational Coefficient and Grey Relational Grade Values}
\scalebox{0.7}{
\begin{tabular}{|p{2.6cm} p{1.9cm} p{2.1cm} p{1.9cm} p{2.1cm}p{1.9cm} p{1.1cm}|  }
\hline
Run No.& \multicolumn{2}{c}{\multirow{1}{*}{Evaluation of $\Delta_{0i}$ }}&\multicolumn{2}{c}{Grey relational coefficient} &GRG&	Rank\\
\hline
& Completeness&	Generalization&	Completeness&	Generalization&		&\\
\hline
Ideal sequence& 1& 1 &1 &1&&\\
\hline
1&	0.661	&0.000&	0.431&	1.000&		0.715&	5\\
2	&0.000	&0.250&	1.000&	0.667&		0.833	&1\\
3&	0.766	&0.313&	0.395	&0.615	&	0.505&	10\\
4&	0.946	&0.750&	0.346&	0.400	&	0.373&	14\\
5&	0.911&	0.875&	0.354&	0.364	&	0.359&	15\\
6&	0.286&	0.187&	0.636&	0.727&		0.682&	6\\
7&	0.446	&0.562&	0.528	&0.471	&	0.499	&11\\
8&	0.411&	0.125&	0.549&	0.800	&	0.675 &	7\\
9&	0.679 &	0.687 &	0.424&	0.421&		0.423 &	12\\
10	&0.625	&0.062&	0.444&	0.889&		0.667&	8\\
11	&0.036 &	0.313&	0.933&	0.615	&	0.774&	2\\
12&	0.554&	0.375&	0.475&	0.571&		0.523&	9\\
13	&0.214	&0.125&	0.700&	0.800	&	0.750&	3\\
14 &	0.821&	1.000&	0.378&	0.333	&	0.356	&16\\
15	 &1.000	&0.500	&0.333&	0.500&		0.417&	13\\
16 &0.054&	0.438&	0.903&	0.533&		0.718&	4\\
\hline
\end{tabular}}
\label{table:grey2}
%\end{sidewaystable}
\end{table}

\begin{figure*}[ht!]
\centering
\includegraphics[width=10.5cm, height=5.7cm]{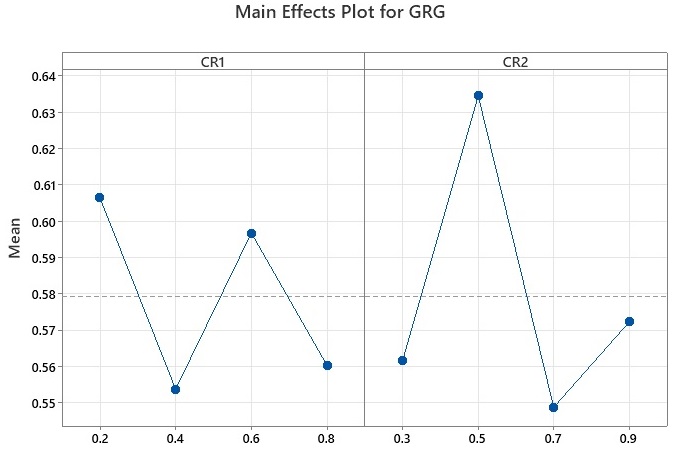}
\caption{ Main effect plot of grey relational grade}
\label{fig:grey4}
\end{figure*} 

\subsection{Parameter Tuning} \label{par}
To find values of the crossover probabilities, $CR_1$ and $CR_2$ suitable for the domain of process discovery, the grey relational analysis combined with the Taguchi approach is used \citep{panda2016multi,lin2004use}. Taguchi method efficiently determines optimal settings of numerous process variables with a minimal set of experiments. Taguchi method suggests replication of the experiment to achieve improved accuracy of the results. Taguchi L16 orthogonal array (OA) design containing 16 experimental runs is used. The results for completeness and generalization are shown in Table~\ref{table:orthogonaltable} and Figure~\ref{figure:O:1:gg1}. Dr. Taguchi's Signal-to-Noise ratios (S$/$N), which are log functions of the desired output, serve as objective functions for optimization \citep{freddi2019introduction}. The optimization of numerous performance variables requires a comprehensive assessment of the S/N ratio. The grey relational analysis is used in the study to solve this issue \citep{lin2004use}. \par

In the grey relational analysis combined with the Taguchi approach, the experimental data is normalized using Equation~\ref{equ:PT1} to avoid different units and to reduce the variability as presented in Table~\ref{table:grey1}. \par 

\begin{equation}
x^{*}_{i}(k)=\dfrac{x_{i}(k)-min(x_{i}(k))}{max (x_{i}(k))- min (x_{i}(k))}
\label{equ:PT1}
\end{equation}

where, i = 1,\ldots, m; k = 1,\ldots, n, m is the number of experimental data and n is the number of responses. $x_{i}$(k) denotes the original value of $k^{th}$ response for $i^{th}$ experimental run, $x^{*}_{i}$(k) denotes the normalized value after the data pre-processing, max ($x_{i}$(k)) denotes the largest value of $x_{i}$(k), min ($x_{i}$(k)) denotes the smallest value of $x_{i}$(k). The next step is to calculate the grey relational coefficient, $\xi_{i}$(k), from the normalized values by using the following equation (Table~\ref{table:grey2}): \par 

\begin{equation}
\xi_{i}(k)=\dfrac{\Delta_{min}- \xi \Delta_{max}}{\Delta_{0i}{k}- \xi \Delta_{max}}
\label{equ:PT2}
\end{equation}

where $\Delta_{0i}$ is the deviation value obtained from the reference value ($x_{0}$(k)) and the comparability value ($x_{i}$(k)). \par

\begin{equation*}
\Delta_{0i}= ||x_{0}(k)-x_{i}(k)||
\end{equation*}

$\Delta_{min}$ and $\Delta_{max}$ are the minimum and maximum values of the absolute difference ($\Delta_{0i}$). $\xi$ is the distinguishing coefficient, where $\xi$ $\in$ [0,1] and value 0.5 is used for experimentation \citep{panda2016multi}. The next step is to find out the grey relational grade (GRG) using the following equation (Table~\ref{table:grey2}): \par 

\begin{equation}
\gamma_{i}=\dfrac{1}{n} \sum_{\mathclap{k=1}}^{n} \xi_{i}(k)
\end{equation}

where $\gamma_{i}$ is the required grey relational grade for the $i^{th}$ experiment. The results are utilized for optimizing the multi-responses as they are converted to a single grade. \par

From the value of GRG, the effects of each process parameter at different levels are plotted and shown in Figure~\ref{fig:grey4}. Using these results optimal settings for the parameters $CR_1$ and $CR_2$ are derived as 0.2 and 0.5 respectively. \par

\begin{table}
\centering
\caption{ Quality dimensions of non-dominated solutions obtained from MoD-ProM algorithm  for synthetic datasets ($f_C$: Completeness, $f_P$: Preciseness, $f_S$: Simplicity, $f_G$: Generalization)}
\scalebox{0.7}{
\begin{tabular}{| p{0.6cm}p{0.6cm}p{0.6cm}p{0.6cm}| p{0.6cm}p{0.6cm} p{0.6cm} p{0.6cm}| p{0.6cm} p{0.6cm} p{0.1cm}p{0.6cm}| p{0.6cm}p{0.6cm} p{0.6cm}p{0.6cm}|p{0.6cm}p{0.6cm}p{0.1cm}p{0.6cm}|}
\hline
\multicolumn{4}{|c|}{\textbf{ETM}}& \multicolumn{4}{c|}{\textbf{g2}}& \multicolumn{4}{c|}{\textbf{g3}} & \multicolumn{4}{c|}{\textbf{g4}}& \multicolumn{4}{c|}{\textbf{g5}}\\	
\hline
\textbf{$f_C$}&	\textbf{$f_G$}&\textbf{$f_S$}&	\textbf{$f_P$}&\textbf{$f_C$}&	\textbf{$f_G$}&	\textbf{$f_S$}&	\textbf{$f_P$}&	\textbf{$f_C$}&	\textbf{$f_G$}&	\textbf{$f_S$}&	\textbf{$f_P$}&	\textbf{$f_C$}&	\textbf{$f_G$}&	\textbf{$f_S$}	&    \textbf{ $f_P$}&\textbf       {$f_C$}&	\textbf{$f_G$}&	\textbf{$f_S$}	& \textbf{$f_P$}\\
\hline
0.625&	0.961&	0.967&	0.435&		0.896&	0.946&	0.996&	0.824&		0.818&	0.97&	1&	0.85&		0.927&	0.969&	0.998&	0.867&		0.916&	0.9695&	1&	0.873\\
	
0.813&	0.9419&	0.984&	0.621&		0.954&	0.934&	0.998&	0.915&		0.906&	0.96&	1&	0.854&		0.947&	0.957&	0.998&	0.916&		0.946&	0.946&	1&	0.916\\	

0.965&	0.911&	0.993&	0.814&		0.972&	0.929&	0.998&	0.941&		0.927&	0.958&	1&	0.872&		0.948&	0.936&	0.999&	0.944&		0.974&	0.942&	1&	0.953\\	

1&	0.79&	0.994&	1&		0.988&	0.923&	0.999&	0.979&		0.947&	0.956&	1&	0.893&		0.96&	0.932&	1&	0.953&		1&	0.938&	1&	1\\

&	&&&		1&	0.916&	1&	1&		0.966&	0.953&	1&	0.915&		0.971&	0.927&	1&	0.96&		&	&	&	\\

&	&	&	&		&	&	&	&		0.978&	0.951&	1&	0.956&		0.982&	0.923&	1&	0.972&		&	&	&	\\
	
&	&	&	&		&	&	&	&		0.99&	0.948&	1&	0.978&		0.988&	0.917&	1&	0.984&		&	&	&	\\	

&	&	&	&		&	&	&	&		1&	0.945&	1&	1&		0.994&	0.911&	1&	0.992&		&	&	&	\\	

&	&	&	&		&	&	&	&		&	&	&	&		1&	0.904&	1&	1&		&	&	&	\\	
\hline	
\multicolumn{4}{|c|}{\textbf{g6}}& \multicolumn{4}{c|}{\textbf{g7}}& \multicolumn{4}{c|}{\textbf{g8}} & \multicolumn{4}{c|}{\textbf{g9}}& \multicolumn{4}{c|}{\textbf{g10}}\\		
\hline
\textbf{$f_C$}&	\textbf{$f_G$}&\textbf{$f_S$}&	\textbf{$f_P$}&\textbf{$f_C$}&	\textbf{$f_G$}&	\textbf{$f_S$}&	\textbf{$f_P$}&	\textbf{$f_C$}&	\textbf{$f_G$}&	\textbf{$f_S$}&	\textbf{$f_P$}&	\textbf{$f_C$}&	\textbf{$f_G$}&	\textbf{$f_S$}	&    \textbf{$f_P$}&\textbf       {$f_C$}&	\textbf{$f_G$}&	\textbf{$f_S$}	& \textbf{$f_P$}\\
\hline
0.621&	0.9623&	1&	0.685&		0.92&	0.979&	1&	0.902&		0.933&	0.945&	1&	0.893&		0.842&	0.947&	1&	0.754&		0.697&	0.938&	1&	0.767\\

0.691&	0.959&	1&	0.667&		0.947&	0.977&	1&	0.945&		0.948&	0.935&	1&	0.932&		0.871&	0.942&	1&	0.795&		0.787&	0.936&	1&	0.835\\

0.781&	0.951&	1&	0.743&		0.967&	0.965&	1&	0.975&		0.96&	0.93&	1&	0.943&		0.907&	0.939&	1&	0.83&		0.872&	0.933&	1&	0.86\\

0.844&	0.948&	1&	0.772&		0.979&	0.963&	1&	0.981&		0.968&	0.925&	1&	0.963&		0.937&	0.936&	1&	0.83&		0.937&	0.932&	1&	0.799\\

0.895&	0.947&	1&	0.766&		0.989&	0.95&	1&	0.994&		0.976&	0.919&	1&	0.983&		0.965&	0.933&	1&	0.829&		0.959&	0.927&	1&	0.875\\

0.928&	0.943&	1&	0.798&		1&	0.947&	1&	1&		0.984&	0.914&	1&	0.987&		0.984&	0.928&	1&	0.945&		0.976&	0.921&	1&	0.935\\

0.96&	0.94&	1&	0.906&		&	&	&	&		0.989&	0.907&	1&	0.989&		1&	0.924&	1&	1&		0.988&	0.914&	1&	0.967\\

0.973&	0.934&	1&	0.945&		&	&	&	&		0.995&	0.9&	1&	0.993&		&	&	&	&		1&	0.907&	1&	1\\

0.986&	0.929&	1&	0.993&		&	&	&	&		0.999&	0.893&	1&	1&		&	&	&	&		&	&	&	\\
1&	0.923&	1&	1&		&	&	&	&		&	&	&	&		&	&	&	&		&	&	&	\\
\hline	
\end{tabular}}
\label{table:DEnon}
\end{table}

\begin{table}
\footnotesize
\caption{ Quality dimensions of non-dominated solutions obtained by NSGA-II in process mining domain  for synthetic datasets ($f_C$: Completeness, $f_P$: Preciseness, $f_S$: Simplicity, $f_G$: Generalization)}
\begin{center}
\scalebox{0.7}{
\begin{tabular}{| p{0.6cm}p{0.6cm}p{0.6cm}p{0.6cm}| p{0.6cm}p{0.6cm} p{0.6cm} p{0.6cm}| p{0.6cm} p{0.6cm} p{0.6cm}p{0.6cm}| p{0.6cm}p{0.6cm} p{0.6cm}p{0.6cm}|p{0.6cm}p{0.6cm}p{0.6cm}p{0.6cm}|}
\hline
\multicolumn{4}{|c|}{\textbf{ETM}}& \multicolumn{4}{c|}{\textbf{g2}}& \multicolumn{4}{c|}{\textbf{g3}} & \multicolumn{4}{c|}{\textbf{g4}}& \multicolumn{4}{c|}{\textbf{g5}}\\	
\hline
\textbf{$f_C$}&	\textbf{$f_G$}&\textbf{$f_S$}&	\textbf{$f_P$}&\textbf{$f_C$}&	\textbf{$f_G$}&	\textbf{$f_S$}&	\textbf{$f_P$}&	\textbf{$f_C$}&	\textbf{$f_G$}&	\textbf{$f_S$}&	\textbf{$f_P$}&	\textbf{$f_C$}&	\textbf{$f_G$}&	\textbf{$f_S$}	&    \textbf{ $f_P$}&\textbf       {$f_C$}&	\textbf{$f_G$}&	\textbf{$f_S$}	& \textbf{ $f_P$}\\
\hline
0.505&	0.957&	0.948&	0.371&		0.515&	0.947&	0.987&	0.497&	0.514&	0.969&	0.996&	0.515& 0.48&	0.97&	0.984&	0.372&	0.529&	0.965&	0.986&	0.525\\
	
0.67&	0.942&	0.959&	0.371&		0.775&	0.947&	0.991&	0.671& 0.698&	0.966&	1&	0.663&	0.532&	0.97&	0.978&	0.396&		0.643&	0.964&	0.992&	0.673\\	

0.83&	0.927&	0.984&	0.628&		0.858&	0.94&	0.994&	0.786& 0.791&	0.963&	0.998&	0.716&	0.568&	0.969&	0.982&	0.412&	0.755&	0.959&	0.996&	0.753\\
	
0.965&	0.911&	0.993&	0.814&		0.936&	0.933&	0.997&	0.888&0.863&	0.956&	1&	0.799&0.666&	0.963&	0.985&	0.532	&		0.809&	0.956&	0.822&	0.796\\	

1&	0.79&	0.994&	1&		0.97&	0.928&	0.998&	0.953&0.899&	0.956&	1&	0.845&	0.774&	0.956&	0.992&	0.674&		0.862&	0.953&	0.999&	0.836\\
	
&	&	&	&		1&	0.916&	1&	1&0.95&	0.952&	1&	0.862&	0.856&	0.943&	0.994&	0.733&		0.892&	0.949&	1&	0.879\\
	
&	&	&	&		&	&	&&0.956&	0.949&	1&	0.87&	0.893&	0.943&	0.996&	0.784&		1&	0.938&	1&	1\\
	
&	&	&	&		&	&	&&0.98&	0.947&	1&	0.91&0.915&	0.937&	0.997&	0.857		&	&	&\\
	
&	&	&	&		&	&	&&0.995&	0.945&	1&	0.966&	0.931&	0.929&	0.998&	0.926		&	&	&\\
\hline	
\multicolumn{4}{|c|}{\textbf{g6}}& \multicolumn{4}{c|}{\textbf{g7}}& \multicolumn{4}{c|}{\textbf{g8}} & \multicolumn{4}{c|}{\textbf{g9}}& \multicolumn{4}{c|}{\textbf{g10}}\\		
\hline
\textbf{$f_C$}&	\textbf{$f_G$}&\textbf{$f_S$}&	\textbf{$f_P$}&\textbf{$f_C$}&	\textbf{$f_G$}&	\textbf{$f_S$}&	\textbf{$f_P$}&	\textbf{$f_C$}&	\textbf{$f_G$}&	\textbf{$f_S$}&	\textbf{$f_P$}&	\textbf{$f_C$}&	\textbf{$f_G$}&	\textbf{$f_S$}	&    \textbf{$f_P$}&\textbf       {$f_C$}&	\textbf{$f_G$}&	\textbf{$f_S$}	& \textbf{ $f_P$}\\
\hline
0.654&	0.961&	0.997&	0.504	&		0.36&	0.982&	0.989&	0.433&		0.634&	0.949&	0.993&	0.59&		0.565&	0.949&	1&	0.548&		0.714&	0.931&	1&	0.583\\

0.712&	0.954&	1&	0.576	&		0.464&	0.98&	0.989&	0.468&		0.666&	0.945&	0.994&	0.604&		0.599&	0.948&	1&	0.641&		0.781&	0.925&	1&	0.535\\

0.765&	0.95&	1&	0.647	&0.559&	0.979&	0.994&	0.577&		0.792&	0.941&	0.996&	0.677&		0.661&	0.946&	1&	0.555&		0.868&	0.925&	1&	0.647\\

0.799&	0.946&	1&	0.72&		0.601&	0.977&	0.992&	0.588&		0.876&	0.939&	0.997&	0.759&		0.693&	0.945&	1&	0.561&		0.898&	0.922&	1&	0.647\\

0.884&	0.94&	1&	0.721	&		0.698&	0.977&	0.996&	0.678&		0.89&	0.926&	0.998&	0.802&		0.726&	0.943&	1&	0.643 &		0.916&	0.917&	1&	0.676\\

0.926&	0.937&	1&	0.751	&		0.704&	0.974&	0.996&	0.642&		0.912&	0.918&	0.999&	0.819&		0.787&	0.94&	1&	0.61 &		0.917&	0.914&	1&	0.739\\

0.978&	0.928&	1&	0.831	&		0.745&	0.97&	0.996&	0.672&		0.965&	0.903&	1&	0.92&		0.846&	0.934&	1&	0.717&		0.942&	0.911&	1&	0.667\\

0.994&	0.923&	1&	0.871	&		0.896&	0.958&	1&	0.854&		0.989&	0.899&	1&	0.945&		0.924&	0.929&	1&	0.754 &		0.971&	0.905&	1&	0.788\\

&&&	&		0.937&	0.954&	1&	0.872&		0.995&	0.892&	1&	0.945&		0.966&	0.927&	1&	0.787 &		0.976&	0.905&	1&	0.805\\
&	&&	&	&	&&&	&		&	&	&	0.984&	0.923&	1&	0.89&		&	&	&	\\

\hline	
\end{tabular}}
\end{center}
\label{table:NSGAnon}
\end{table}

\begin{table}
\centering
\caption{ Quality dimensions of non-dominated solutions obtained by NSGA-II for process mining and MoD-ProM algorithm for real-life datasets  ($f_C$: Completeness, $f_P$: Preciseness, $f_S$: Simplicity, $f_G$: Generalization)}
\scalebox{0.7}{
\begin{tabular}{| p{0.7cm}p{0.7cm}p{0.7cm}p{0.8cm}| p{0.7cm}p{0.7cm} p{0.7cm} p{0.8cm}| p{0.7cm} p{0.7cm} p{0.7cm}p{0.78cm}| p{0.7cm}p{0.7cm} p{0.7cm}p{0.75cm}|}
\hline	
\multicolumn{16}{|c|}{\textbf{Non-dominated solutions obtained from NSGA-II for process mining algorithm}}\\
\hline
\multicolumn{4}{|c|}{\textbf{BPI 2012}}& \multicolumn{4}{c|}{\textbf{BPI 2013}}& \multicolumn{4}{c|}{\textbf{BPI 2018}} & \multicolumn{4}{c|}{\textbf{Sepsis}}\\	
\hline
\textbf{$f_C$}&	\textbf{$f_G$}&\textbf{$f_S$}&	\textbf{$f_P$}&\textbf{$f_C$}&	\textbf{$f_G$}&	\textbf{$f_S$}&	\textbf{$f_P$}&	\textbf{$f_C$}&	\textbf{$f_G$}&	\textbf{$f_S$}&	\textbf{$f_P$}&	\textbf{$f_C$}&	\textbf{$f_G$}&	\textbf{$f_S$}	&    \textbf{$f_P$}\\
\hline
0.794	&0.9839&	0.9984&	0.6945	&		0.6637&	0.9834&	0.9964&	0.6956&		0.9671&	0.9857&	0.9461	&0.406&		0.7509&	0.9529	&0.9955&	0.499\\	
0.815&	0.9838&	0.9986&0.6756	&		0.8453 &	0.9758	&0.9953 &	0.6433 &		0.9873	&0.9824&	0.9568&	0.5989&		0.8198	&0.9524	&0.9958&	0.538\\	
0.892&	0.9836&	0.9987&0.6819	&		0.9749&	0.975	&0.997&	0.7879&		0.9886	&0.9792&	0.9848&	0.7978&		0.8872	&0.9448	&0.9979&	0.664\\	
0.899&	0.9835&	0.9989&	0.7151	&		0.9816 &	0.9749	&0.9971&	0.805&		0.99	&0.979 &	0.9768&	0.7922&		0.9033	&0.943	&0.9974&	0.649\\	
0.907&	0.9834&	0.999&	0.6688	&		0.9832&	0.9699&	0.9973&	0.8308&		0.9998	&0.9236	&0.9907&	0.9855&		0.9158	&0.9345	&0.9981&	0.704\\
0.914&	0.9832&	0.9991&	0.6874	&		0.9871	&0.9685&	0.9985&0.8818&		&	&	&			&		0.9165	&0.931	&0.9977&	0.670\\
0.929&	0.9831&	0.9992&	0.7271&		0.9916&	0.9683&	0.9988&	0.8915&		&	&	&			&	0.9434	&0.9299	&0.998 &	0.714\\
0.931&	0.982&	0.9993&	0.7056	&		0.9966	&0.9241&	0.9983&	0.869&		&	&	&			&		&	&	&\\
&	&	&			&	0.9974&	0.9239&	0.9987&	0.8458	&	&	&			&&		&	&	&\\
\hline	
\multicolumn{16}{|c|}{\textbf{Non-dominated solutions obtained from MoD-ProM algorithm}}\\
\hline	
\multicolumn{4}{|c|}{\textbf{BPI 2012}}& \multicolumn{4}{c|}{\textbf{BPI 2013}}& \multicolumn{4}{c|}{\textbf{BPI 2018}} & \multicolumn{4}{c|}{\textbf{Sepsis}}\\	
\hline
\textbf{$f_C$}&	\textbf{$f_G$}&\textbf{$f_S$}&	\textbf{$f_P$}&\textbf{$f_C$}&	\textbf{$f_G$}&	\textbf{$f_S$}&	\textbf{$f_P$}&	\textbf{$f_C$}&	\textbf{$f_G$}&	\textbf{$f_S$}&	\textbf{$f_P$}&	\textbf{$f_C$}&	\textbf{$f_G$}&	\textbf{$f_S$}	&    \textbf{$f_P$}\\
\hline
0.9146&0.9853	&0.9995&0.9229&		0.9942	&0.9768 &0.998&	0.7398	&		0.9965	&0.9793	&0.985&0.806&		0.994	&0.961	&0.9994&	0.849\\

0.9957&0.985	&0.9996	&0.9294	&		0.9946 &	0.97 &	0.9979&0.8078 &		0.9986&	0.9792&	0.9848&	0.7978&		0.9957&	0.949&	0.9995&	0.892\\	

0.9965&0.9849	&0.9997	&0.930	&		0.9948 &	0.9696&	0.9985&	0.8312&		0.999&	0.979&	0.981&0.792&		0.9973	&0.936&	0.9997&	0.941\\	

0.9974&0.9828	&0.9998	&0.9310	&		0.996&	0.9687&	0.9989&	0.8548&		0.9998&	0.9236	&0.993&	0.9858&		0.9977&0.908	&0.9998&	0.972\\
	
&	&	&	&		0.9961	&0.9686&	0.999&	0.8543&		&	&&&		&&&\\

&	&	&	&	0.9981 &	0.9618 &	0.9996&0.9524&		&&&&		&&&\\

&	&	&	&	0.9987&	0.9617&	0.9994&0.8672&		&	&	&&		&&&\\

&	&&&		0.9988 &	0.9245&	0.9995&0.8791&		&	&	&	&		&&&\\
\hline
\end{tabular}}
\label{table:DEnon1}
\end{table}

\begin{figure*}[!ht]
\captionsetup{font=footnotesize}
           \centering
\subcaptionbox{}{\includegraphics[width=0.32\textwidth,height=4cm]{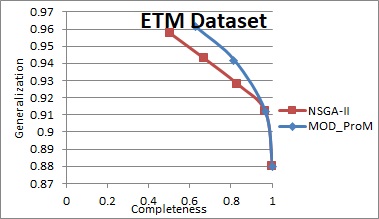}}\label{figure:O:1:ga-2}%
\hfill % <-- Seperation
\subcaptionbox{}{\includegraphics[width=0.32\textwidth,height=4cm]{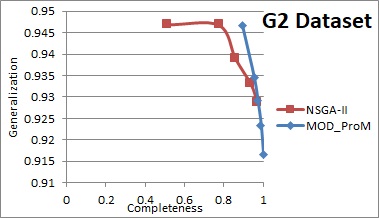}} \label{figure:O:1:gb-1}%
\hfill % <-- Seperation
\subcaptionbox{}{\includegraphics[width=0.32\textwidth,height=4cm]{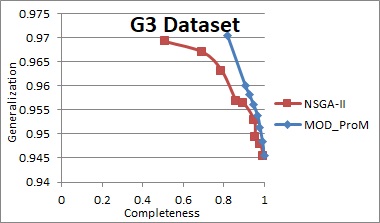}}\label{figure:O:1:gc}%
\hfill % <-- Seperation
\subcaptionbox{}{\includegraphics[width=0.32\textwidth,height=4cm]{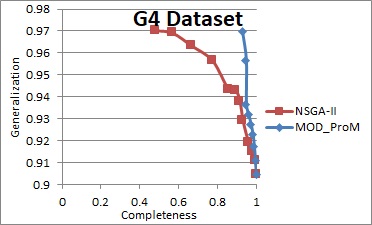}}\label{figure:O:1:gd}%
\hfill % <-- Seperation
\subcaptionbox{}{\includegraphics[width=0.32\textwidth,height=4cm]{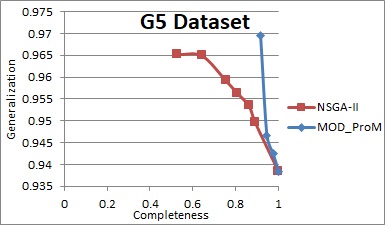}}\label{figure:O:1:ge}%
\hfill % <-- Seperation
\subcaptionbox{}{\includegraphics[width=0.32\textwidth,height=4cm]{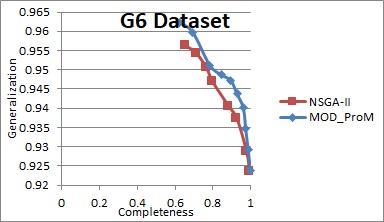}}\label{figure:O:1:gf}%
\hfill % <-- Seperation
\subcaptionbox{}{\includegraphics[width=0.32\textwidth,height=4cm]{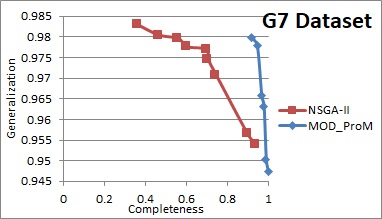}}\label{figure:O:1:gg}%
\hfill % <-- Seperation
\subcaptionbox{}{\includegraphics[width=0.32\textwidth,height=4cm]{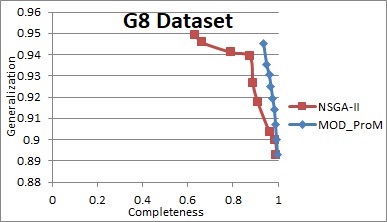}}\label{figure:O:1:gh}%
\hfill % <-- Seperation
\subcaptionbox{}{\includegraphics[width=0.32\textwidth,height=4cm]{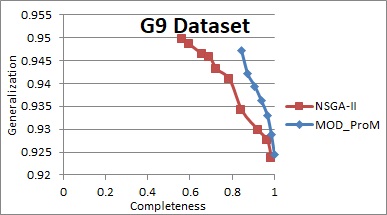}}\label{figure:O:1:gi}%
\hfill % <-- Seperation
\subcaptionbox{}{\includegraphics[width=0.32\textwidth,height=4cm]{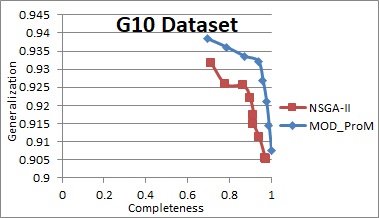}}\label{figure:O:1:gj}%
\caption{Pareto-curve for the non-dominated solutions for NSGA-II and MoD-ProM for Synthetic datasets.}
 \label{figure:O:1:gg2}
        \end{figure*}
        
\begin{figure*}[!ht]
\captionsetup{font=footnotesize}
           \centering
\subcaptionbox{}{\includegraphics[width=0.32\textwidth,height=4cm]{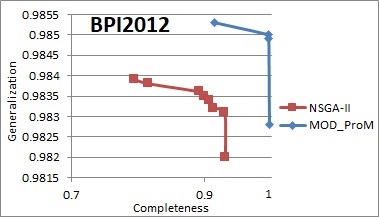}}\label{figure:O:1:ha}%
\hfill % <-- Seperation
\subcaptionbox{}{\includegraphics[width=0.32\textwidth,height=4cm]{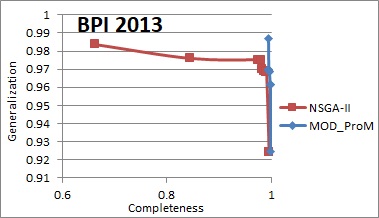}} \label{figure:O:1:hb}%
\hfill % <-- Seperation
\subcaptionbox{}{\includegraphics[width=0.32\textwidth,height=4cm]{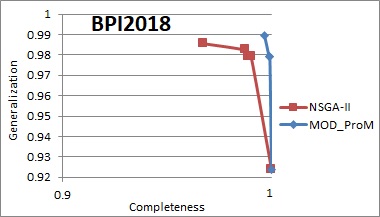}}\label{figure:O:1:hc}%
\hfill % <-- Seperation
\subcaptionbox{}{\includegraphics[width=0.32\textwidth,height=4cm]{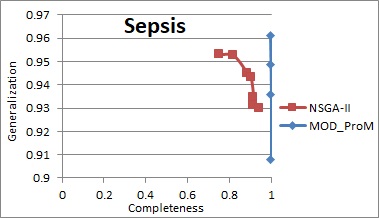}}\label{figure:O:1:hd}%
\hfill % <-- Seperation
\caption{Pareto-curve for the non-dominated solutions for NSGA-II and MoD-ProM for real-life datasets.}
 \label{figure:O:1:hg2}
        \end{figure*}

\begin{table*}[t]
\caption{ Quality dimensions of the models using the state-of-the-art algorithms ($f_C$: Completeness, $f_P$: Preciseness, $f_S$: Simplicity, $f_G$: Generalization)}
\begin{center}
\scalebox{0.7}{
\begin{tabular}{| p{1.5cm}|p{0.2cm}|p{0.7cm}p{0.6cm} p{0.6cm} p{0.6cm}p{0.6cm} p{0.6cm}p{0.6cm} p{0.6cm} p{0.6cm}p{0.6cm}|}
\hline
&&	\textbf{ETM}&\textbf{g2}&\textbf{g3}&\textbf{g4}&	\textbf{g5}&\textbf{g6}&\textbf{g7}&\textbf{g8}&\textbf{g9}&\textbf{g10}\\
\hline
\multirow{4}{*}{\begin{minipage}{0.5in}\textbf{Genetic Miner}\end{minipage}}&$f_C$&0.3&	1&	0.31&	0.59&	1	&1&	1&	0.26&	0.48&	0.48\\

	&$f_P$&	0.94&	1	&0.6	&1	&1&	1	&1	&0.15	&1&	1\\

	&$f_S$	&1	&1&	1&	0.97&	1&	1& 1&	0.72&	0.96&	0.88\\

	&$f_G$&	0.56&	0.91&	0.88	&0.90	&0.921	&0.80	&0.91	&0.88&0.75	&0.61\\
\hline
\multirow{4}{*}{\begin{minipage}{0.5in}\textbf{Heuristic Miner}\end{minipage}}	&$f_C$&	0.37&	1&	1	&0.78&	1&	0.66&	1&	0.52&	0.74&	0.78\\

	&$f_P$&0.98	&1	&1&	1&	1	&0.99	&1	&1&	1	&1\\
	
	&$f_S$&1&	1&	1&	1&	1&	0.99&	0.98&	0.93&	0.96&	1\\
	
	&$f_G$	&0.62&	0.913&	0.89&	0.81&	0.92&	0.80&	0.81&	0.90&	0.73&	0.60\\
	\hline
\multirow{4}{*}{\begin{minipage}{0.5in}\textbf{\(\alpha\)$^{++}$}\end{minipage}}&	$f_C$&	0.89&	0.33&	0&	1&	1&	0.45&	0&	0.35&	0.48&	0.563\\

&	$f_P$&	1&	0.96&	0.18&	0.97&	1	&1&	0.12&	1&	1&	1\\

&	$f_S$&	1&	0.78&	0.79&	1	&1&	0.76	&0.93&	0.74&	0.79&	0.76\\

&	$f_G$&	0.56&	0.62&	0.74&	0.91&	0.92&	0.84&	0.81&	0.91&	0.59&	0.43\\
	\hline
\multirow{4}{*}{\begin{minipage}{0.5in}\textbf{ILP}\end{minipage}}&	$f_C$&	1	&1&	1&	1&	1&	1	&1&	1&1&	1\\

&	$f_P$	&1	&0.97&	0.97	&1	&1&	0.99	&1&	0.98&	0.98&	0.95\\

&	$f_S$	&0.93&	0.93&	0.92&	0.96&	1	&0.74&	0.93&	0.667	&0.9	&0.68\\

&	$f_G$	&0.79&	0.99	&0.93&	0.91&	0.921&	0.79&	0.91&	0.92&	0.76&	0.61\\
	\hline
	\multirow{4}{*}{\begin{minipage}{0.5in}\textbf{Inductive Miner}\end{minipage}}&	$f_C$&	0.89	&0.958&	0.757&	0.70&	0.80&	0.63	&0.74&0.79	&0.668&0.61	\\

&	$f_P$	&1	&0.89&	0.73	&0.56	&0.75&	0.41	&0.64&	0.637&0.423	&	0.26\\

&	$f_S$	&1&	0.9&	0.82&	0.81&	0.9	&0.7&0.85	&0.63	&0.84	&0.65\\

&	$f_G$	&0.56&	0.91	&0.94&	0.91&	0.94&0.91	&0.95	&0.91	&0.88	&0.9	\\
	\hline
\end{tabular}}
\end{center}
\label{table:singleobjective}
\end{table*}
     
\begin{figure*}[!ht]
\captionsetup{font=footnotesize}
           \centering
\subcaptionbox{}{\includegraphics[width=0.32\textwidth,height=4cm]{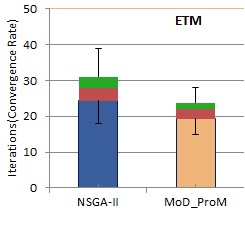}}\label{figure:O:1:ja}%
\hfill % <-- Seperation
\subcaptionbox{}{\includegraphics[width=0.32\textwidth,height=4cm]{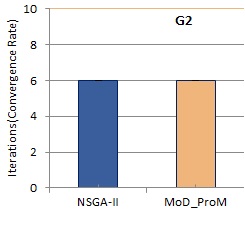}} \label{figure:O:1:jb}%
\hfill % <-- Seperation
\subcaptionbox{}{\includegraphics[width=0.32\textwidth,height=4cm]{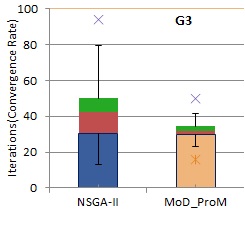}}\label{figure:O:1:jc}%
\hfill % <-- Seperation
\subcaptionbox{}{\includegraphics[width=0.32\textwidth,height=4cm]{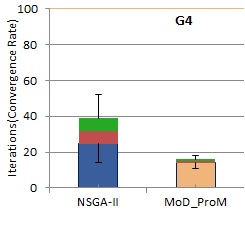}}\label{figure:O:1:jd}%
\hfill % <-- Seperation
\subcaptionbox{}{\includegraphics[width=0.32\textwidth,height=4cm]{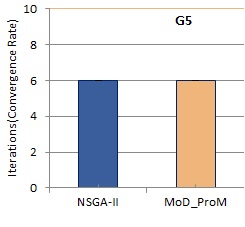}}\label{figure:O:1:je}%
\hfill % <-- Seperation
\subcaptionbox{}{\includegraphics[width=0.32\textwidth,height=4cm]{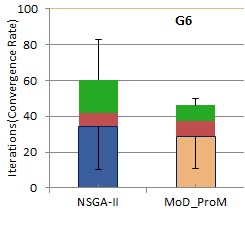}}\label{figure:O:1:jf}%
\hfill % <-- Seperation
\subcaptionbox{}{\includegraphics[width=0.32\textwidth,height=4cm]{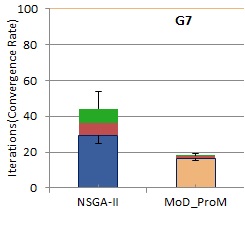}}\label{figure:O:1:jg}%
\hfill % <-- Seperation
\subcaptionbox{}{\includegraphics[width=0.32\textwidth,height=4cm]{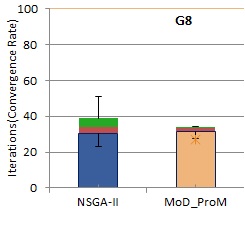}}\label{figure:O:1:jh}%
\hfill % <-- Seperation
\subcaptionbox{}{\includegraphics[width=0.32\textwidth,height=4cm]{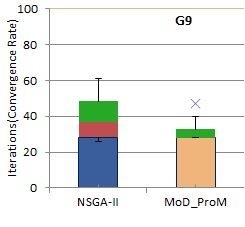}}\label{figure:O:1:ji}%
\hfill % <-- Seperation
\subcaptionbox{}{\includegraphics[width=0.32\textwidth,height=4cm]{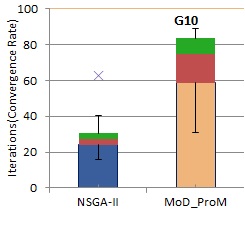}}\label{figure:O:1:jj}%
\caption{Convergence rate for NSGA-II and MoD-ProM for synthetic datasets.}
 \label{figure:O:1:jg2}
        \end{figure*}

\begin{figure*}[!ht]
\captionsetup{font=footnotesize}
           \centering
\subcaptionbox{}{\includegraphics[width=0.32\textwidth,height=4cm]{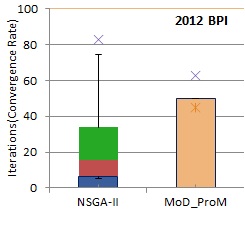}}\label{figure:O:1:ka}%
\hfill % <-- Seperation
\subcaptionbox{}{\includegraphics[width=0.32\textwidth,height=4cm]{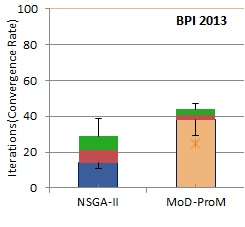}} \label{figure:O:1:kb}%
\hfill % <-- Seperation
\subcaptionbox{}{\includegraphics[width=0.32\textwidth,height=4cm]{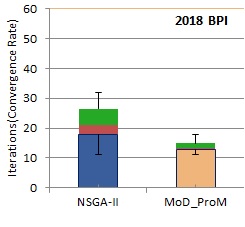}}\label{figure:O:1:kc}%
\hfill % <-- Seperation
\subcaptionbox{}{\includegraphics[width=0.32\textwidth,height=4cm]{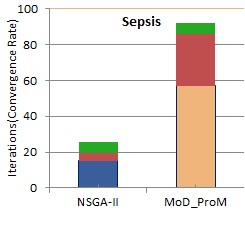}}\label{figure:O:1:kd}%
\hfill % <-- Seperation
\caption{Convergence rate for NSGA-II and MoD-ProM for real-life datasets.}
 \label{figure:O:1:kg2}
        \end{figure*}
        
\begin{figure*}[!ht]
\captionsetup{font=footnotesize}
\centering
\subcaptionbox{Synthetic Dataset}{\includegraphics[width=\textwidth,height=4cm]{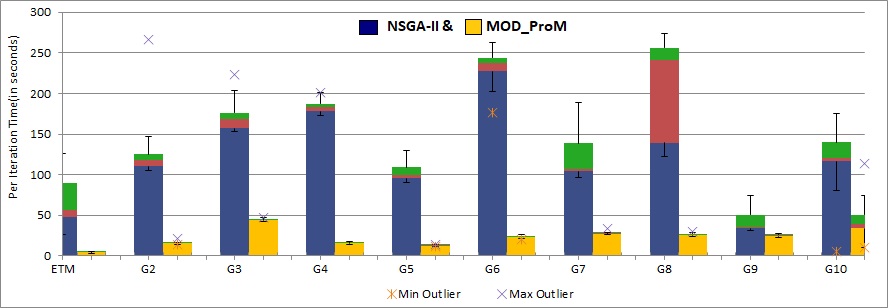}}\label{figure:O:1:sa}%
\hfill % <-- Seperation
\subcaptionbox{Real-Life Dataset}{\includegraphics[width=0.58\textwidth,height=4cm]{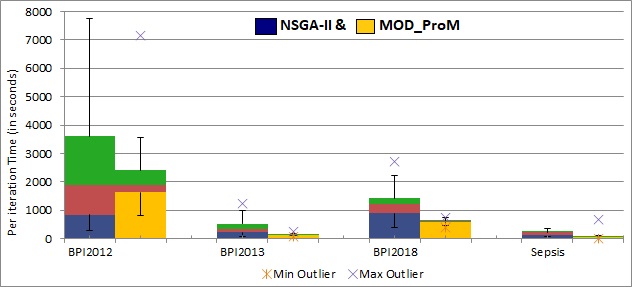}} \label{figure:O:1:sb}%
\caption{Comparison of per iteration running time for NSGA-II/MoD-ProM, over 30 runs.}
 \label{figure:O:1:sg2}
        \end{figure*}
         
\subsection{Analysis of the Results} \label{sec:results}
The proposed algorithm (MoD-ProM) is run for the real-life and for the synthetic datasets and the values for quality dimensions, namely completeness ($f_c$), preciseness ($f_p$), simplicity ($f_s$), and generalization ($f_g$), for the discovered non-dominated solutions are shown in Tables ~\ref{table:DEnon} and ~\ref{table:DEnon1}, respectively. \par

The proposed approach is compared with the NSGA-II algorithm for process discovery. Tables ~\ref{table:NSGAnon} and ~\ref{table:DEnon1} present the values for the quality dimensions for the discovered non-dominated solutions for real-life and synthetic datasets, respectively. \par

Pareto-curves for the non-dominated solutions of NSGA-II and the proposed multi-objective differential evolution for process mining (MoD-ProM) are plotted for comparison (Figure ~\ref{figure:O:1:gg2} and ~\ref{figure:O:1:hg2}). The Pareto-curves show that in 12 out of 14 datasets, the results of the proposed algorithm are superior to the NSGA-II algorithm.\par 

We also compute the convergence rate and per iteration computation time for NSGA-II and the proposed MoD-ProM, over 30 runs (Figures ~\ref{figure:O:1:jg2}, ~\ref{figure:O:1:kg2}, and ~\ref{figure:O:1:sg2}). While in 2 datasets, the algorithms (NSGA-II, MoD-ProM) show a similar convergence rate, in 8 out of 14 datasets, the proposed MoD-ProM converges faster than NSGA-II, demonstrating superior exploration of the proposed approach. Figure ~\ref{figure:O:1:sg2} shows that in all cases, the proposed algorithm is superior to NSGA-II in terms of running time per iteration. It is evident from the results that NSGA-II is computationally more expensive than the proposed MoD-ProM algorithm. \par

The proposed algorithm is also compared with Genetic Miner, Heuristic Miner, \(\alpha\)$^{++}$, ILP, and Inductive Miner. To rank the proposed approach and the traditional algorithms, additional comparison based on a weighted average \citep{buijs2012role} of the quality dimensions is made (Table~\ref{table:Buijs}). \cite{buijs2012role} proposed a weighted average computation methodology suitable to the process mining domain, as follows:  \par 

 \begin{equation} \label{equ:ws}
Weighted\ Sum = (10* f_c+(1*f_p)+(1*f_s)+(1*f_g))/13
\end{equation} 

where for a given process model, $f_c$, $f_p$, $f_s$, and $f_g$ denote the completeness, preciseness, simplicity, and generalization values, respectively. A higher weight is assigned to completeness as the process model should be able to reproduce the behavior expressed in the event log. \par
 
Table~\ref{table:singleobjective} shows the quality dimensions for the process model discovered by the state-of-the-art algorithms. The results (Table~\ref{table:Buijs}) show that the proposed algorithm produces superior-quality process models for all the datasets in terms of the weighted average. \par

It is also observed that the models generated through the optimization of a combination of completeness and generalization exhibit superior values for the other quality dimensions. \par

\begin{table*}[t]
%\footnotesize
\caption{Weighted sum (Equation \ref{equ:ws}) of the quality dimensions \citep{buijs2012role} and the resulting algorithm ranks for each dataset. Rank 7 indicates the best solution}
\begin{center}
\scalebox{0.7}{
\begin{tabular}{| p{0.7cm}|p{0.8cm}p{0.8cm}p{1.2cm} p{0.8cm}p{1.2cm} p{1.0cm} p{1.05cm}| p{0.35cm} p{0.35cm} p{1.2cm}p{0.4cm}p{1.2cm}p{0.8cm} p{0.8cm}|}
\hline
Event Log&\multicolumn{7}{c}{Weighted sum} & \multicolumn{7}{|c|}{ Rank}\\
\hline
&	GM&	HM&	\(\alpha\)$^{++}$&	ILP&Inductive Miner&	NSGA-II&	MoD-ProM&	GM&	HM&	\(\alpha\)$^{++}$&	ILP&Inductive Miner&	NSGA-II&MoD-ProM\\
\hline
ETM&	0.423&	0.485&	0.881&	0.978&0.881&	0.983&	0.983&		1&	2&	3.5&	5&3.5&	6.5&	6.5\\

g2&	0.993&	0.993&	0.435&	0.991&0.94&	0.994&	0.994&		4.5&	4.5&	1&3&	2&	6.5&	6.5\\

g3&	0.429&	0.991&	0.132&	0.986&0.77&	0.989&	0.996&		2&	6&	1&4&	3&	5&	7\\

g4&	0.675&	0.816&	0.991&	0.99&0.71&	0.993&	0.993&		1&	3&	5&4&2&	6.5&	6.5\\

g5&	0.994&	0.994&	0.994&	0.994&0.81&	0.995&	0.995&		3.5&3.5&3.5&3.5&1	&	6.5&	6.5\\

g6&	0.984&	0.722&	0.546&	0.963&0.64&	0.98&	0.994&		6&3&	1&4&2&5&	7\\

g7&	0.993&	0.984&	0.143&	0.988&0.756&	0.99&	0.996&		6&3&	1&4&2&5&	7\\

g8&	0.334&	0.618&	0.473&	0.966&0.795&	0.984&	0.992&		1&	3&	2&5&	4&	6&	7\\

g9&	0.578&	0.776&	0.552&	0.972&0.679&	0.974&	0.994&		2&	4&	1&	5&3&6&7\\

g10&	0.561&	0.8&	0.602&	0.941&0.6&	0.959&	0.993&		1&	4&3&	5&2&	6&7\\

\hline
\end{tabular}}
\end{center}
\label{table:Buijs}
\end{table*}

\begin{figure}[!ht]
%\caption{Nodes and Arcs in a Petri net}
\centering
\includegraphics[width=0.78\textwidth,height=11cm]{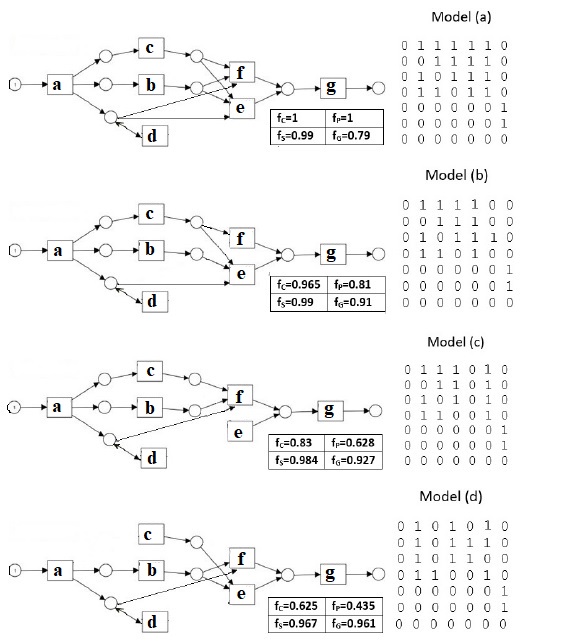}
\caption{Petri Net and causality relation matrix of non-dominated process models of the proposed algorithm for ETM dataset}
\label{fig:petE}
\end{figure}

\begin{figure}[!ht]
%\caption{Nodes and Arcs in a Petri net}
\centering
\includegraphics[width=0.78\textwidth,height=11cm]{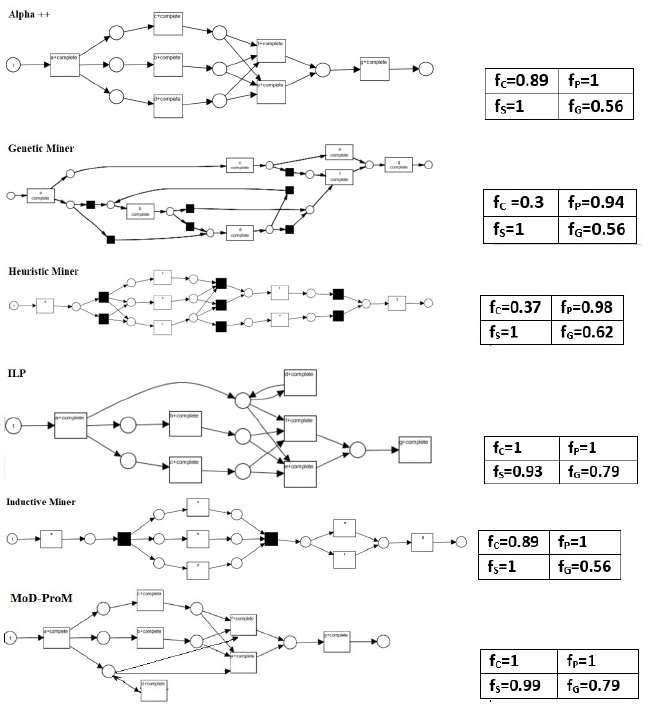}
\caption{Petri Net of the process models discovered by the compared algorithms for ETM dataset}
\label{fig:pet}
\end{figure}

\subsubsection{Process Model Representation} \label{sec:processModelRep}
As discussed earlier (Section \ref{sec:initialization} on Initialization), the proposed approach represents a process model as a causality relation matrix \citep{de2007genetic}. However, many state-of-the-art approaches use other semantics, such as Petri net, BPMN models, DFGs, etc. Petri net is possibly the more popular technique for visualizing the discovered process model. We apply the methodology given by \cite{de2007genetic} to map between a Petri net and a causality relation matrix.\par

To better explain our results, we have graphically depicted the discovered models (causality relation matrices) as Petri nets for the ETM event log. ETM is a popular dataset in the literature comprising seven tasks. Being a small dataset, it is feasible to show (Figure~\ref{fig:petE}) the causality relation matrices and the corresponding Petri nets of the four models discovered by the proposed algorithm (MoD-ProM). \par
 
For the ETM dataset, the ProM tool generated Petri nets for the state-of-the-art algorithms is shown in Figure ~\ref{fig:pet}. To compare with the proposed approach, the Petri net of the model with the highest completeness value, discovered by the proposed MoD-ProM, is also drawn in Figure ~\ref{fig:pet}. \par

The completeness or replay fitness \citep{van2016process} quantifies the ability to replay the trace from an event log onto the Petri net \citep{alves2006genetic}. That is, a process model (Petri net) will exhibit a perfect completeness value if every process instance in the given event log can be replayed (simulated) in the Petri net. For the ETM dataset (Figure ~\ref{fig:pet}), it is observed that the proposed MoD-ProM algorithm and the ILP algorithm can replay every process instance in the event log. It is observed that the process model discovered by Inductive miner and \(\alpha\)$^{++}$ do not replay some of the traces, such as (a,b,c,f,g) and (a,c,d,f,g). Traces (a,b,c,d,e,g) and (a,c,b,d,f,g) are not replayed by the model generated by the heuristic miner algorithm. Similarly, the process model generated by the genetic miner algorithm does not replay (a,c,d,b,f,g) and (a,d,c,b,f,g). \par

\section{{Conclusions and Future Works}} \label{sec:concl}
While conventional process mining algorithms generate a unique process model from the event logs, multi-objective evolutionary algorithms generate several candidate models. The goodness of the generated process models is measured based on quality dimensions such as completeness, generalization, simplicity, and preciseness. A practitioner in the field of process mining may select the most appropriate process model based on the domain requirement. For example, if a user requires a model that replays the maximum number of traces, he/she may pick the model with a better value of completeness \citep{deshmukh2020moea}.\par

In this paper, we are using the idea of differential evolution towards generating a Pareto-front in the domain of process discovery, a first attempt in this direction. The proposed algorithm performs optimization using completeness and generalization as objective functions. These two quality dimensions make a good pair, as a model with high generalization value can help in improving the current system and can be used for designing future improved processes. Completeness is an important quality dimension because a discovered process model is expected to describe the behavior stored in the log. \par

The experiments were run for ten synthetic and four real-life datasets, and are repeated 30 times for each dataset. The results are compared with state-of-the-art process discovery algorithms such as \(\alpha\)$^{++}$, heuristic miner, genetic miner, ILP, and Inductive Miner, and also with NSGA-II for process discovery. \par

Results show that the models generated by the proposed approach vis-a-vis the compared approaches exhibit a higher value for all the quality dimensions indicating the discovery of ‘‘good’’ process models. The non-dominated solutions generated by the proposed approach (MoD-ProM) are better than those generated by the NSGA-II algorithm for process discovery. The Pareto curve shows that the results of the proposed algorithm are superior or at least as good as that of the NSGA-II algorithm. In terms of computational time requirement, the MoD-ProM algorithm performs consistently better for all datasets as compared to the NSGA-II algorithm.\par
 
In summary, we present a novel proposal for process model discovery. The approach employs a multi-objective differential evolution method to optimize the novel combination of completeness and generalization. Results show that the proposed approach is computationally efficient in discovering good-quality process models. {However, the proposed approach is limited by the hardware availability.} \par

%Limitations
{In the future, we plan to evaluate the applicability of recent multi-objective algorithms \citep{ abed2024ibja, alawad2023binary} in the domain of process discovery and study their computational complexity. In addition, to address the computational intensity and time consumption of process discovery for large event logs, we can explore parallel implementations (multi-core processors, GPU-based processing, and distributed computing environments) for the proposed algorithm. } \par 

%By using parallel computing techniques, the efficiency and scalability of process discovery algorithms can be improved. This involves distributing workloads across multiple processors or computing nodes, reducing , in the futureprocessing time, and managing larger datasets more effectively. 

%{ In addition, to address the computational intensity and time consumption of process discovery for large event logs, we plan to explore parallel implementations in process mining. By using parallel computing techniques, the efficiency and scalability of process discovery algorithms can be improved. This involves distributing workloads across multiple processors or computing nodes, reducing processing time, and managing larger datasets more effectively. We will investigate multi-core processors, GPU-based processing, and distributed computing environments to evaluate their performance on real-world event logs. This will aim to develop more robust and scalable process mining tools for timely organizational insights.}

%Since process discovery for large volumes of event logs is a computationally intensive and time-consuming task, as a part of future work, we intend to explore parallel implementations in the domain of process mining.
% MoD\_ProM exhibits lower computational cost as compared to NSGA-II. 

\section{Data availability}
Previously reported data (event logs) were used to support this study and are described in the section \ref{sec:experiment} on Experimentation. These prior studies (and datasets) are cited at relevant places within the text (References \citep{van2012event,steeman2013bpi,challenge20184tu,alves2006genetic,vazquez2015prodigen}).

\section{Conflicts of Interest}
The author(s) declare(s) that there is no conflict of interest regarding the publication of this paper.

\section{Funding Statement}
This research did not receive any specific grant from funding agencies in public, commercial, or not-for-profit sectors.
\clearpage
\bibliographystyle{hindawi_bib_style}           % Style BST file.
\bibliography{bibModProm}

\end{document}